Journal on
Artificial Intelligence

Tech Science Press

**REVIEW**

# A Comprehensive Overview and Comparative Analysis on Deep Learning Models

**Farhad Mortezapour Shiri** [1, *]**, Thinagaran Perumal**[1]**, Norwati Mustapha**[1]**, and Raihani Mohamed**[1]

[1] Faculty of Computer Science and Information Technology, University Putra Malaysia (UPM), Serdang, 43400, Malaysia

*Corresponding Author: Farhad Mortezapour Shiri. Email: GS63904@student.upm.edu.my

**ABSTRACT**

Deep learning (DL) has emerged as a powerful subset of machine learning (ML) and artificial intelligence (AI), outperforming traditional ML methods, especially in handling unstructured and large datasets. Its impact spans across various domains, including speech recognition, healthcare, autonomous vehicles, cybersecurity, predictive analytics, and more. However, the complexity and dynamic nature of real-world problems present challenges in designing effective deep learning models. Consequently, several deep learning models have been developed to address different problems and applications. In this article, we conduct a comprehensive survey of various deep learning models, including Convolutional Neural Network (CNN), Recurrent Neural Network (RNN), Temporal Convolutional Networks (TCN), Transformer, Kolmogorov-Arnold networks (KAN), Generative Models, Deep Reinforcement Learning (DRL), and Deep Transfer Learning. We examine the structure, applications, benefits, and limitations of each model. Furthermore, we perform an analysis using three publicly available datasets: IMDB, ARAS, and Fruit-360. We compared the performance of six renowned deep learning models: CNN, RNN, Long Short-Term Memory (LSTM), Bidirectional LSTM, Gated Recurrent Unit (GRU), and Bidirectional GRU alongside two newer models, TCN and Transformer, using the IMDB and ARAS datasets. Additionally, we evaluated the performance of eight CNN-based models, including VGG (Visual Geometry Group), Inception, ResNet (Residual Network), InceptionResNet, Xception (Extreme Inception), MobileNet, DenseNet (Dense Convolutional Network), and NASNet (Neural Architecture Search Network), for image classification tasks using the Fruit-360 dataset.

## 1 Introduction

Artificial intelligence (AI) aims to emulate human-level intelligence in machines. In computer science, AI refers to the study of "intelligent agents," which are objects capable of perceiving their environment and taking actions to maximize their chances of achieving specific goals [1]. Machine learning (ML) is a field that focuses on the development and application of methods capable of learning from datasets [2]. ML finds extensive use in various domains, such as speech recognition, computer vision, text analysis, video games, medical sciences, and cybersecurity.

In recent years, deep learning (DL) techniques, a subset of machine learning (ML), have outperformed traditional ML approaches across numerous tasks, driven by several critical advancements [3]. The proliferation of large datasets has been pivotal in enabling models to learn intricate patterns and relationships, thereby significantly enhancing their performance [4]. Concurrently, advancements in hardware acceleration technologies, notably Graphics Processing Units (GPUs) and Field-Programmable Gate Arrays (FPGAs) [5] have markedly reduced model training times by facilitating rapid computations and parallel processing capabilities. These advancements have substantially accelerated the training process. Moreover, enhancements in algorithmic techniques for optimization and training have further augmented the speed and efficiency of deep learning models, leading to quicker convergence and superior generalization capabilities [4]. Deep learning techniques have demonstrated remarkable success across a wide range of applications, including computer vision (CV), natural language processing (NLP), and speech recognition. These applications underscore the transformative impact of DL in various domains, where it continues to set new performance benchmarks [6, 7].

Deep learning models draw inspiration from the structure and functionality of the human nervous system and brain. These models employ input, hidden, and output layers to organize processing units. Within each layer, the nodes or units are interconnected with those in the layer below, and each connection is assigned a weight value. The units sum the inputs after multiplying them by their corresponding weights [8]. Fig. 1 illustrates the relationship between AI, ML, and DL, highlighting that machine learning and deep learning are subfields of artificial intelligence.

The objective of this research is to provide a comprehensive overview of various deep learning models and compare their performance across different applications. In Section 2, we introduce a fundamental definition of deep learning. Section 3 covers supervised deep learning models, including Multi-Layer Perceptron (MLP), Convolutional Neural Networks (CNN), Recurrent Neural Networks (RNN), Temporal Convolutional Networks (TCN), and Kolmogorov-Arnold Networks (KAN). Section 4 reviews generative models such as Autoencoders, Generative Adversarial Networks (GANs), and Deep Belief Networks (DBNs). Section 5 presents a comprehensive survey of the Transformer architecture. Deep Reinforcement Learning (DRL) is discussed in Section 6, while Section 7 addresses Deep Transfer Learning (DTL). The principles of hybrid deep learning are explored in Section 8, followed by a discussion of deep learning applications in Section 9. Section 10 surveys the challenges in deep learning and potential alternative solutions. In Section 11, we conduct experiments and analyze the performance of different deep learning models using three datasets. Research directions and future aspects are covered in Section 12. Finally, Section 13 concludes the paper.

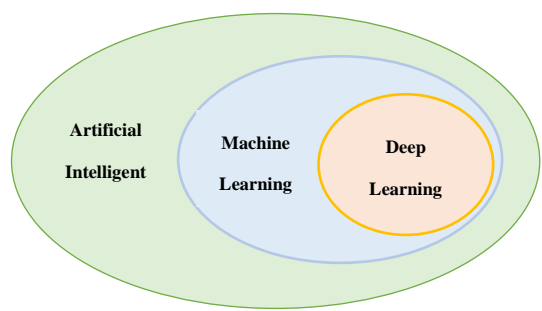

**Figure 1**. Relationship between artificial intelligence (AI), machine learning (ML), and deep learning (DL).

## 2 Deep Learning

Deep learning (DL) involves the process of learning hierarchical representations of data by utilizing architectures with multiple hidden layers. With the advancement of high-performance computing facilities, deep learning techniques using deep neural networks have gained increasing popularity [9]. In a deep learning algorithm, data is passed through multiple layers, with each layer progressively extracting features and transmitting information to the subsequent layer. The initial layers extract low-level characteristics, which are then combined by later layers to form a comprehensive representation [6].

In traditional machine learning techniques, the classification task typically involves a sequential process that includes pre-processing, feature extraction, meticulous feature selection, learning, and classification. The effectiveness of machine learning methods heavily relies on accurate feature selection, as biased feature selection can lead to incorrect class classification. In contrast, deep learning models enable simultaneous learning and classification, eliminating the need for separate steps. This capability makes deep learning particularly advantageous for automating feature learning across diverse tasks [10]. Fig. 2 visually illustrates the distinction between deep learning and traditional machine learning in terms of feature extraction and learning.

In the era of deep learning, a wide array of methods and architectures have been developed. These models can be broadly categorized into two main groups: discriminative (supervised) and generative (unsupervised) approaches. Among the discriminative models, two prominent groups are Convolutional Neural Networks (CNNs) and Recurrent Neural Networks (RNNs). Additionally, generative approaches encompass various models such as Generative Adversarial Networks (GANs) and Auto-Encoders (AEs) [11]. In the following sections, we provide a comprehensive survey of different types of deep learning models.

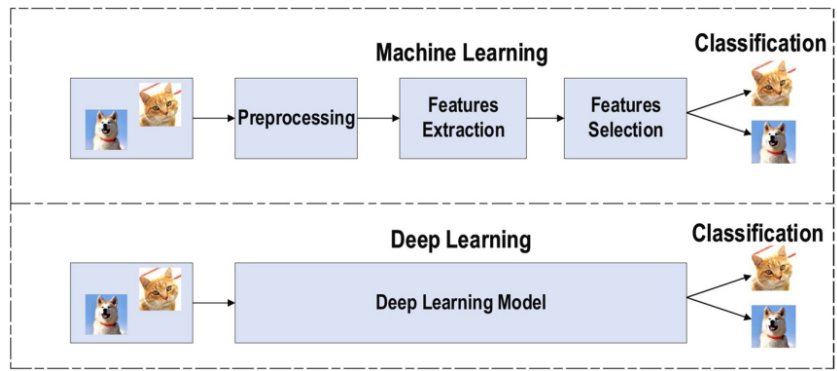

**Figure 2**. Visual illustration of the distinction between deep learning and traditional machine learning in terms of feature extraction and learning [10].

## 3 Supervised Deep Learning Models

In supervised learning and classification tasks, this family of deep learning algorithms is used to perform discriminative functions. These supervised deep architectures typically model the posterior distributions of classes based on observable data, enabling effective pattern classification. Common supervised models include Multi-Layer Perceptron (MLP), Convolutional Neural Networks (CNN), Recurrent Neural Networks (RNN), Temporal Convolutional Networks (TCN), Kolmogorov-Arnold Networks (KAN), and their variations. A brief overview of these methods follows.

### 3.1 Multi Layers Perceptron (MLP)

The Multi-Layer Perceptron (MLP) model is a type of feedforward artificial neural network

(ANN) that serves as a foundation architecture for deep learning or deep neural networks (DNNs) [11]. It operates as a supervised learning approach. The MLP consists of three layers: the input layer, the output layer, and one or more hidden layers [12]. It is a fully connected network, meaning each neuron in one layer is connected to all neurons in the subsequent layer.

In an MLP, the input layer receives the input data and performs feature normalization. The hidden layers, which can vary in number, process the input signals. The output layer makes decisions or predictions based on the processed information [13]. Fig. 3 (a) depicts a single-neuron perceptron model, where the activation function φ (Eq. (1)) is a non-linear function used to map the summation function $(xw + b)$ to the output value $y$.

$$y = \varphi(xw + b) \tag{1}$$

In Eq. (1), the terms $x, w, b$, and $y$ represent the input vector, weighting vector, bias, and output value, respectively [14]. Fig. 3 (b) illustrates the structure of the multilayer perceptron (MLP) model.

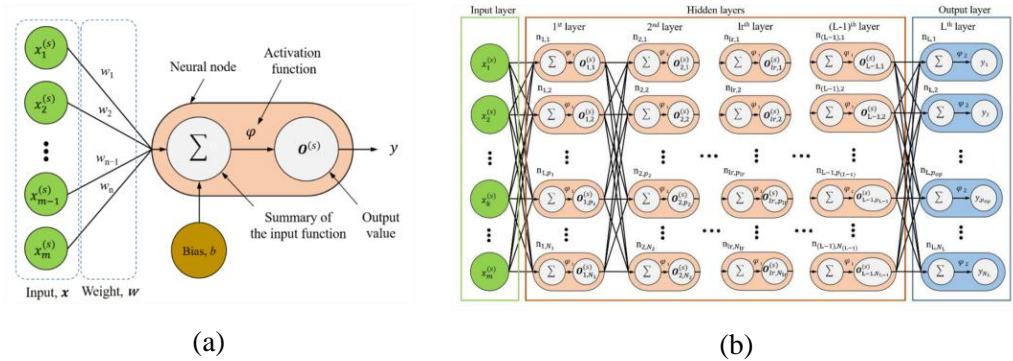

(a)                                                          (b)

**Figure 3**. (a) Single-neuron perceptron model. (b) Structure of the MLP [14].

### 3.2 Convolutional Neural Networks (CNN)

Convolutional Neural Networks (CNNs) are a powerful class of deep learning models widely applied in various tasks, including object detection, speech recognition, computer vision, image classification, and bioinformatics [15]. They have also demonstrated success in time series prediction tasks [16]. CNNs are feedforward neural networks that leverage convolutional structures to extract features from data [17]. CNN has a two-stage architecture that combines a classifier and a feature extractor to provide automatic feature extraction and end-to-end training with the least amount of pre-processing necessary [18]. Unlike traditional methods, CNNs automatically learn and recognize features from the data without the need for manual feature extraction by humans [19]. The design of CNNs is inspired by visual perception [17]. The major components of CNNs include the convolutional layer, pooling layer, fully connected layer, and activation function [20, 21]. Fig. 4 presents the pipeline of the convolutional neural network, highlighting how each layer contributes to the efficient processing and successful progression of input data through the network.

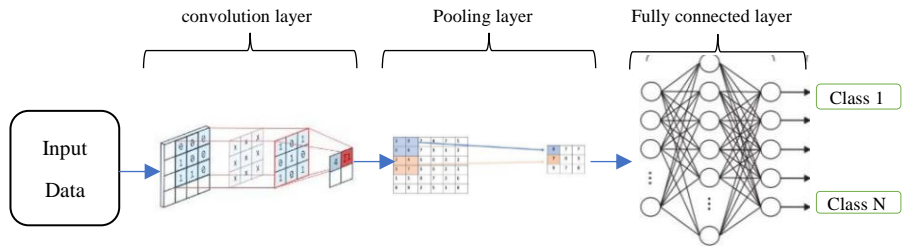

**Figure 4.** The pipeline of a Convolutional Neural Network.

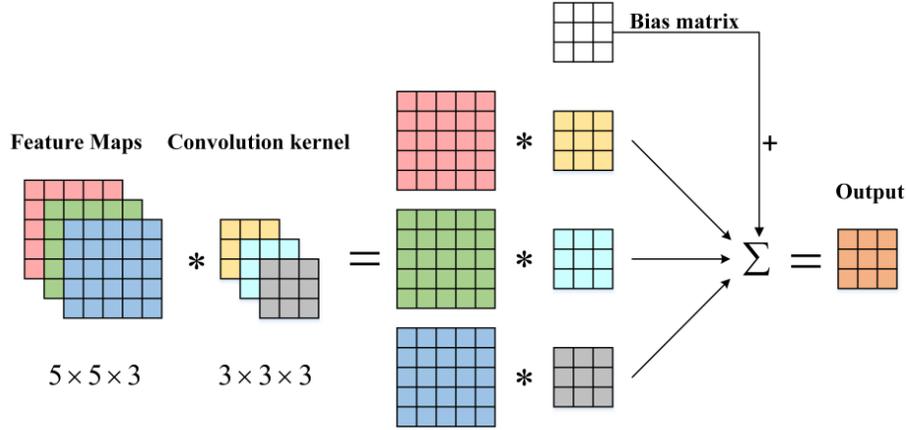

**Figure 5.** Schematic diagram of the convolution process [22].

**Convolutional Layer:** The convolutional layer is a pivotal component of CNN. Through multiple convolutional layers, the convolution operation extracts distinct features from the input. In image classification, lower layers tend to capture basic features such as texture, lines, and edges, while higher layers extract more abstract features. The convolutional layer comprises learnable convolution kernels, which are weight matrices typically of equal length, width, and an odd number (e.g., 3x3, 5x5, or 7x7). These kernels are convolved with the input feature maps, sliding over the regions of the feature map and executing convolution operations [22]. Fig. 5 illustrates the schematic diagram of the convolution process.

**Pooling Layer:** Typically following the convolutional layer, the pooling layer reduces the number of connections in the network by performing down-sampling and dimensionality reduction on the input data [23]. Its primary purpose is to alleviate the computational burden and address overfitting issues [24]. Moreover, the pooling layer enables CNN to recognize objects even when their shapes are distorted or viewed from different angles, by incorporating various dimensions of an image through pooling [25]. The pooling operation produces output feature maps that are more robust against distortion and errors in individual neurons [26]. There are various pooling methods, including Max Pooling, Average Pooling, Spatial Pyramid Pooling, Mixed Pooling, Multi-Scale Order-Less, and Stochastic Pooling [27-30]. Fig. 6 depicts an example of Max Pooling, where a window slides across the input, and the contents of the window are processed by a pooling function [31].

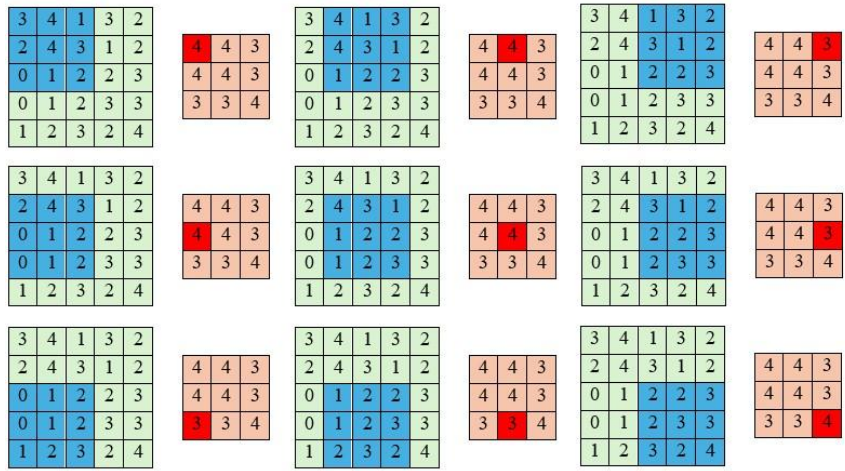

**Figure 6.** Computing the output values of a $3 \times 3$ max pooling operation on a $5 \times 5$ input.

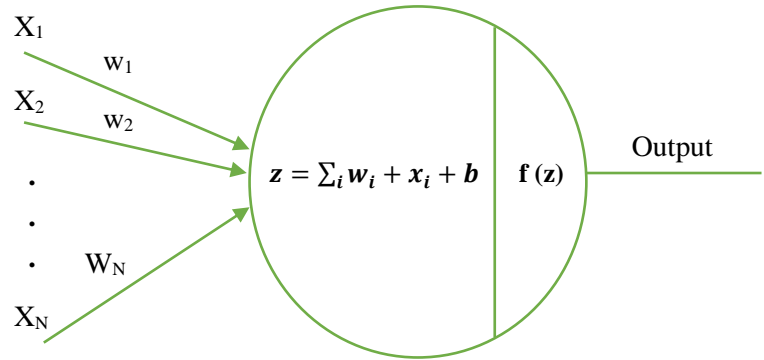

**Figure 7.** The general structure of activation functions.

**Fully Connected (FC) Layer:** The FC layer is typically located at the end of a CNN architecture. In this layer, every neuron is connected to all neurons in the preceding layer, adhering to the principles of a conventional multi-layer perceptron neural network. The FC layer receives input from the last pooling or convolutional layer, which is a vector created by flattening the feature maps. The FC layer serves as the classifier in the CNN, enabling the network to make predictions [10].

**Activation Functions**: Activation functions are fundamental components in convolutional neural networks (CNNs), indispensable for introducing non-linearity into the network. This non-linearity is crucial for CNN's ability to model complex patterns and relationships within the data, allowing it to perform tasks beyond simple linear classification or regression. Without non-linear activation functions, a CNN would be limited to linear operations, significantly constraining its capacity to accurately represent the intricate, non-linear behaviors typical of many real-world phenomena [32].

Fig. 7 typically illustrates how these activation functions modulate input signals to produce output, highlighting the non-linear transformations applied to the input data across different regions of the function curve. In this figure, $x_i$ represents the input feature, while $w_{ij}$ denotes the weight associated with the connection between the input feature $x_i$ and neuron $j$. The figure shows that neuron $j$ receives $n$ features simultaneously. The output from neuron $j$ is labeled by $y_j$, and its internal state, or bias, is indicated by $b_j$. The activation function, depicted as $f(.)$, could be any one of several types such as the Rectified Linear Unit (ReLU), hyperbolic tangent (Tanh), Sigmoid function, or others [33, 34].

These various activation functions are shown in Fig. 8, with emphasis on their distinct characteristics and profiles. These activation functions are essential for convolutional neural networks (CNNs) to be more effective in a variety of applications by allowing them to recognize intricate patterns and provide accurate predictions. Sigmoid and Tanh functions are frequently referred to as saturating nonlinearities due to the way they act when inputs are very large or small. As per the reference, the Sigmoid function approaches values of 0 or 1, whereas the Tanh function leans towards -1 or 1[17]. Different alternative nonlinearities have been suggested for reducing problems associated with these saturating effects, including Rectified Linear Unit (ReLU) [35], Leaky ReLU [36], Parametric Rectified Linear Units (PReLU) [37], Randomized Leaky ReLU (RReLU) [38], S-shaped ReLU (SReLU) [39], and Exponential Linear Units (ELUs) [40], Gaussian Error Linear Units (GELUs) [41].

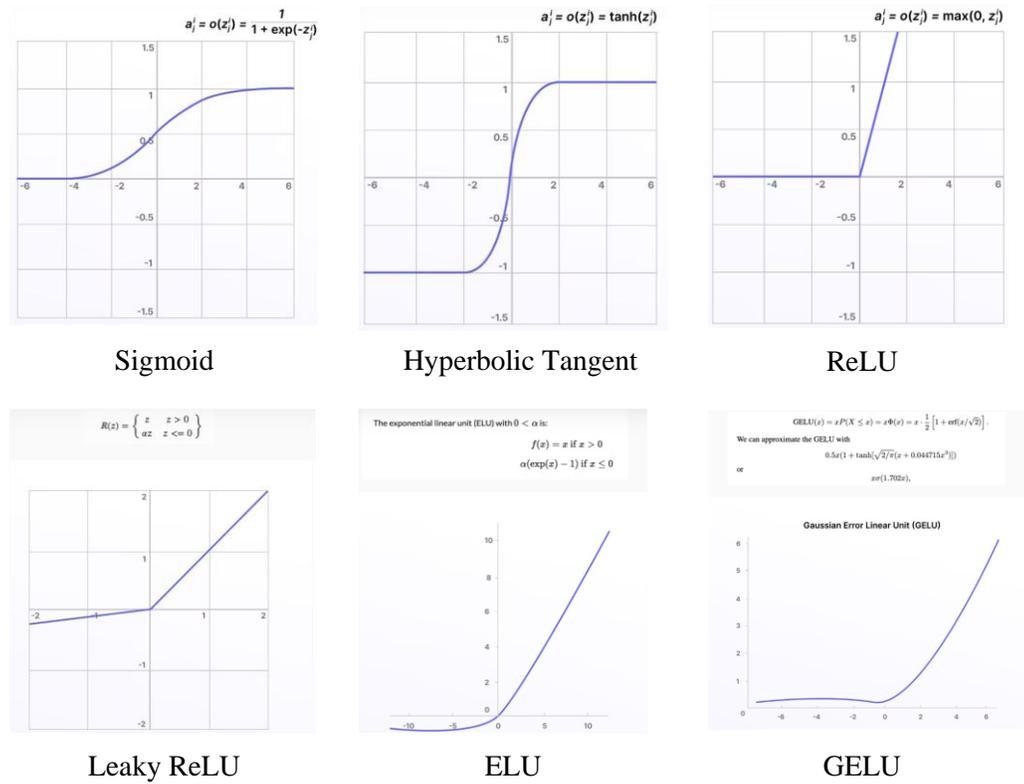

**Figure 8.** Diagram of different activation functions.

ReLU (Rectified Linear Unit) is one of the most often used activation functions in modern CNNs because of how well it solves the vanishing gradient issue during training. The definition of ReLU in mathematics is as Eq. (2), where the input to the neuron is represented by $x$ [34].

$$f(x) = \max(0, x) = \begin{cases} x_i, & if \ x_i \geq 0 \\ 0, & if \ x_i < 0 \end{cases} \tag{2}$$

This feature helps CNN learn complicated features more efficiently by effectively "turning off" any negative input values while maintaining positive values. It also keeps neurons from being saturated during training.

As an alternative, the definition of the Sigmoid function is represented by Eq. (3), where $x$ stands for the input of the neuron.

$$f(x) = \frac{1}{e^{-x}} \tag{3}$$

Although the sigmoid distinctive S-shape and capacity to condense real numbers into a range between 0 and 1 make it useful for binary classification, its propensity to saturate can hinder training by causing the vanishing gradient problem in deep neural networks.

Convolutional Neural Networks (CNNs) are extensively used in various fields, including natural language processing, image segmentation, image analysis, video analysis, and more. Several CNN variations have been developed, such as AlexNet [42], VGG (Visual Geometry Group) [43], Inception [44, 45], ResNet (Residual Networks) [46, 47], WideResNet [48], FractalNet [49], SqueezeNet [50], InceptionResNet [51], Xception (Extreme Inception) [52], MobileNet [53, 54], DenseNet (Dense Convolutional Network) [55], SENet (Squeeze-and-Excitation Network) [56], Efficientnet [57, 58] among others. These variants are applied in different application areas based on their learning capabilities and performance.

### 3.3 Recurrent Neural Networks (RNN)

Recurrent Neural Networks (RNNs) are a class of deep learning models that possess internal memory, enabling them to capture sequential dependencies. Unlike traditional neural networks that treat inputs as independent entities, RNNs consider the temporal order of inputs, making them suitable for tasks involving sequential information [59]. By employing a loop, RNNs apply the same operation to each element in a series, with the current computation depending on both the current input and the previous computations [60].

The ability of RNNs to utilize contextual information is particularly valuable in tasks such as natural language processing, video classification, and speech recognition. For example, in language modeling, understanding the preceding words in a sentence is crucial for predicting the next word. RNNs excel at capturing such dependencies due to their recurrent nature[61-63].

However, a limitation of simple RNN is their short-term memory, which restricts their ability to retain information over long sequences [64]. To overcome this, more advanced RNN variants have been developed, including Long Short-Term Memory (LSTM) [65], bidirectional LSTM [66], Gated Recurrent Unit (GRU) [67], bidirectional GRU [68], Bayesian RNN [69], and others.

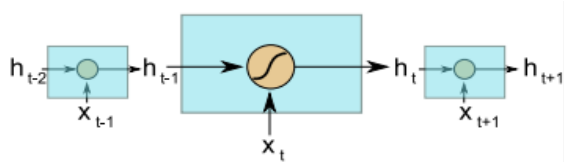

**Figure 9.** Simple RNN internal operation [70].

Fig. 9 depicts a simple recurrent neural network, where the internal memory $(h_t)$ is computed using Eq. (4) [70]:

$$h_t = g(Wx_t + Uh_t + b) \tag{4}$$

In this equation, $g()$ represents the activation function (typically Tanh), $U$ and $W$ are adjustable weight matrices for the hidden state $(h)$, $b$ is the bias, and $x$ denotes the input vector.

RNNs have proven to be powerful models for processing sequential data, leveraging their ability to capture dependencies over time. The various types of RNN models, such as LSTM, bidirectional LSTM, GRU, and bidirectional GRU, have been developed to address specific challenges in different applications.

### 3.3.1 Long Short-Term Memory (LSTM)

Long Short-Term Memory (LSTM) is an advanced variant of Recurrent Neural Networks (RNN) that addresses the issue of capturing long-term dependencies. LSTM was initially introduced by [65] in 1997 and further improved by [71] in 2013, gaining significant popularity in the deep learning community. Compared to standard RNN, LSTM models have proven to be more effective at retaining and utilizing information over longer sequences.

In an LSTM network, the current input at a specific time step and the output from the previous time step are fed into the LSTM unit, which then generates an output that is passed to the next time step. The final hidden layer of the last time step, sometimes along with all hidden layers, is commonly employed for classification purposes [72]. The overall architecture of an LSTM network is depicted in Fig. 10 (a). LSTM consists of three gates: input gate, forget gate, and output gate. Each gate performs a specific function in controlling the flow of information. The input gate decides how to update the internal state based on the current input and the previous internal state. The forget gate determines how much of the previous internal state should be forgotten. Finally, the output gate regulates the influence of the internal state on the system [60, 73].

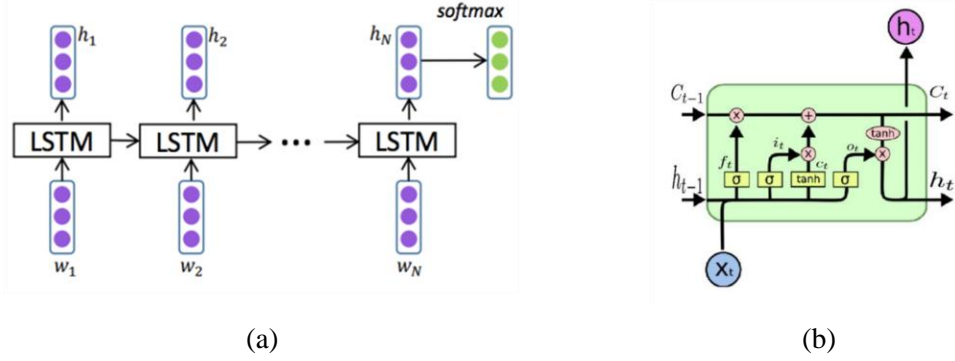

(a)                                                                                    (b)

**Figure 10**. (a) The high-level architecture of LSTM. (b) The inner structure of LSTM unit [60].

Fig. 10 (b) illustrates the update mechanism within the inner structure of an LSTM. The update for the LSTM unit is expressed by Eq. (5):

$$
\begin{cases}
h^{(t)} = g_o^{(t)} f_h\big(s^{(t)}\big) \\
s^{(t-1)} = g_f^{(t)} s^{(t-1)} + g_i^{(t)} f_s\big(wh^{(t-1)}\big) + uX^{(t)} + b \\
g_i^{(t)} = sigmoid\,(w_i h^{(t-1)} + u_i X^{(t)} + b_i) \\
g_f^{(t)} = sigmoid\,(w_f h^{(t-1)} + u_f X^{(t)} + b_f) \\
g_o^{(t)} = sigmoid\,(w_o h^{(t-1)} + u_o X^{(t)} + b_o)
\end{cases}
\tag{5}
$$

where $f_h$ and $f_s$ represent the activation functions of the system state and internal state, typically utilizing the hyperbolic tangent function. The gating operation, denoted as g, is a feedforward neural network with a sigmoid activation function, ensuring output values within the range of [0, 1], which are interpreted as a set of weights. The subscripts $i, o,$ and $f$ correspond to the input gate, output gate, and forget gate, respectively.

While standard LSTM has demonstrated promising performance in various tasks, it may struggle to comprehend input structures that are more complex than a sequential format. To address this limitation, a tree-structured LSTM network, known as S-LSTM, was proposed by [74]. S-LSTM consists of memory blocks comprising an input gate, two forget gates, a cell gate, and an output gate. While S-LSTM exhibits superior performance in challenging sequential modeling problems, it comes with higher computational complexity compared to standard LSTM [75].

### 3.3.2 Bidirectional LSTM

Bidirectional Long Short-Term Memory (Bi-LSTM) is an extension of the LSTM architecture that addresses the limitation of standard LSTM models by considering both past and future context in sequence modeling tasks. While traditional LSTM models process input data only in the forward direction, Bi-LSTM overcomes this limitation by training the model in two directions: forward and backward [76, 77].

A Bi-LSTM consists of two parallel LSTM layers: one processes the input sequence in the forward direction, while the other processes it in the backward direction. The forward LSTM layer reads the input data from left to right, as indicated by the green arrow in Fig. 11. Simultaneously, the backward LSTM layer reads the input data from right to left, as represented by the red arrow [78]. This bidirectional processing enables the model to capture information from both past and future contexts, allowing for a more comprehensive understanding of temporal dependencies within the sequence.

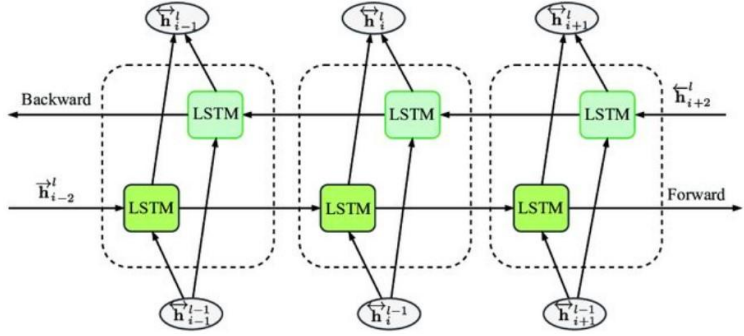

**Figure 11.** The architecture of a Bidirectional LSTM model [76].

During the training phase, the forward and backward LSTM layers independently extract features and update their internal states based on the input sequence. The output of each LSTM layer at each time step is a prediction score. These prediction scores are then combined using a weighted sum to generate the final output result [78]. By incorporating information from both directions, Bi-LSTM models can capture a broader context and improve the model's ability to model temporal dependencies in sequential data.

Bi-LSTM has been widely applied in various sequence modeling tasks such as natural language processing, speech recognition, and sentiment analysis. It has shown promising results in capturing complex patterns and dependencies in sequential data, making it a popular choice for tasks that require an understanding of both past and future context.

### 3.3.3 Gated Recurrent Unit (GRU)

The Gated Recurrent Unit (GRU) is another variant of the RNN architecture that addresses the short-term memory issue and offers a simpler structure compared to LSTM [59]. GRU combines the input gate and forget gate of LSTM into a single update gate, resulting in a more streamlined design. Unlike LSTM, GRU does not include a separate cell state. A GRU unit consists of three main components: an update gate, a reset gate, and the current memory content. These gates enable the GRU to selectively update and utilize information from previous time steps, allowing it to capture long-term dependencies in sequences [79]. Fig. 12 illustrates the structure of a GRU unit [80].

The update gate (Eq. (6)) determines how much of the past information should be retained and combined with the current input at a specific time step. It is computed based on the concatenation of the previous hidden state $h_{t-1}$ and the current input $x_t$, followed by a linear transformation and a sigmoid activation function.

$$z_t = \sigma(W_z[h_{t-1}, x_t] + b_z) \tag{6}$$

The reset gate (Eq. (7)) decides how much of the past information should be forgotten. It is computed in a similar manner to the update gate using the concatenation of the previous hidden state and the current input.

$$r_t = \sigma(W_r[h_{t-1}, x_t] + b_r) \tag{7}$$

The current memory content (Eq. (8)) is calculated based on the reset gate and the concatenation of the transformed previous hidden state and the current input. The result is passed through a hyperbolic tangent activation function to produce the candidate activation.

$$\tilde{h}_t = tanh(W_h[r_t h_{t-1}, x_t]) \tag{8}$$

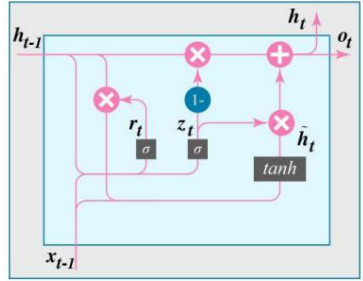

**Figure 12.** The structure of a GRU unit [80].

Finally, the final memory state $h_t$ is determined by a combination of the previous hidden state and the candidate activation (Eq. (9)). The update gate determines the balance between the previous hidden state and the candidate activation. Additionally, an output gate $o_t$ can be introduced to control the information flow from the current memory content to the output (Eq. (10)). The output gate is computed using the current memory state $h_t$ and is typically followed by an activation function, such as the sigmoid function.

$$h_t = (1 - z_t)h_{t-1} + z_t \tilde{h}_t \tag{9}$$

$$o_t = \sigma_o(W_o h_t + b_o) \tag{10}$$

where the weight matrix of the output layer is $W_o$ and the bias vector of the output layer is $b_o$.

GRU offers a simpler alternative to LSTM with fewer tensor operations, allowing for faster training. However, the choice between GRU and LSTM depends on the specific use case and problem at hand. Both architectures have their advantages and disadvantages, and their performance may vary depending on the nature of the task [59].

### 3.3.4 Bidirectional GRU

The Bidirectional Gated Recurrent Unit (Bi-GRU) [81] improves upon the conventional GRU architecture through the integration of contexts from the past and future in sequential modeling tasks. In contrast to the conventional GRU, which exclusively processes input sequences forward, the Bi-GRU manages sequences in both forward and backward directions. In order to do this, two parallel GRU layers are used, one of which processes the input data forward and the other in reverse [82]. Fig. 13 shows the Bi-GRU's structural layout.

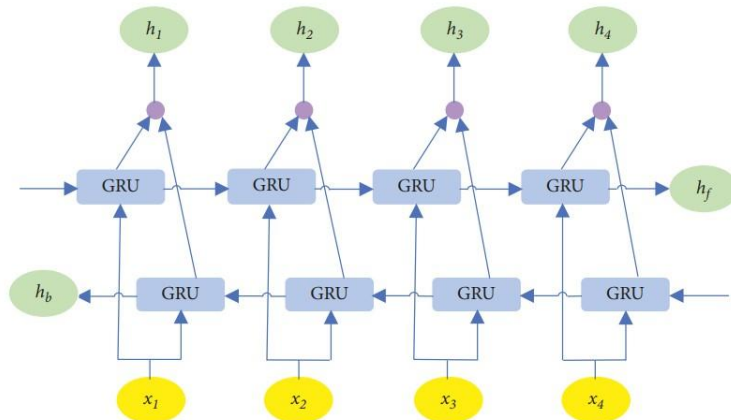

**Figure 13.** The structure of a Bi-GRU model [83].

### *3.4 Temporal Convolutional Networks (TCN)*

Temporal Convolutional Networks (TCN) represent a significant advancement in neural network architectures, specifically tailored for handling sequential data, particularly time series. Originating as an extension of the one-dimensional Convolutional Neural Network (CNN), TCN was first introduced by [84] in 2017 for the task of action segmentation in video data, and its application was further generalized to other types of sequential data by [85] in 2018. TCN retains the powerful feature extraction capabilities inherent to CNN while being highly efficient in processing and analyzing time series data.

The purpose of training TCN is to forecast the next $l$ values of the input time series. Assume that we have a sequence of inputs $x_0, x_1, \ldots, x_l$. We would like to predict, at each time step, some corresponding output $y_0, y_1, \ldots, y_l$, whose values are equal to the inputs shifted forward $l$ time steps. The primary limitation is that it can only use the inputs that have already been observed: $x_0, x_1, \ldots, x_t$, when forecasting the output $y_t$ for a given time step $t$ [86]. TCN is characterized by two fundamental properties: (1) The convolutions within the network are causal, ensuring that the output at any given time step depends solely on the current and past inputs, without any influence from future inputs. (2) Similar to Recurrent Neural Networks (RNNs), TCN can process sequences of arbitrary length and produce output sequences of identical length. The three primary components of a typical TCN are residual connections, dilated convolution, and causal convolution [85, 87, 88]. Fig. 14 illustrates the schematic architecture of a TCN model.

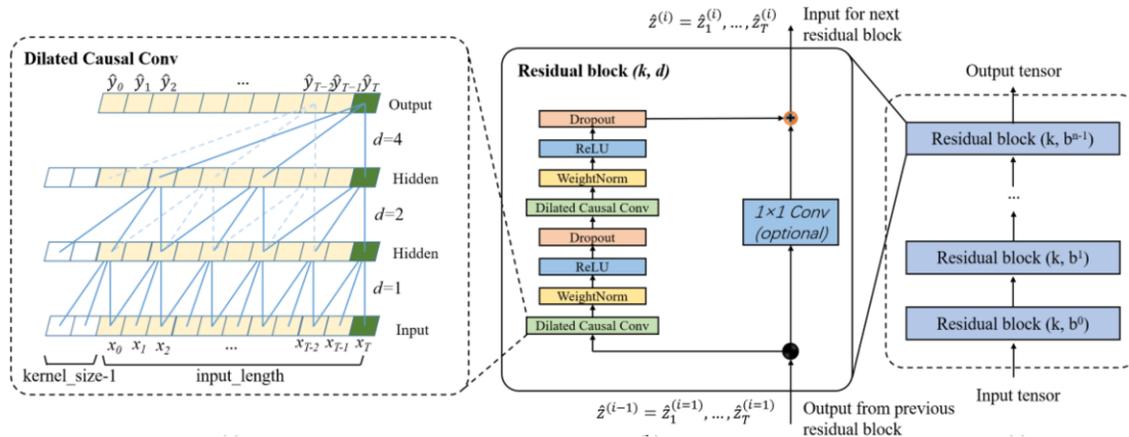

**Figure 14.** Schematic diagram of the TCN model architecture [89].

### Causal Convolution:

The TCN architecture is built upon two foundational principles. To adhere to the first principle, the initial layer of a TCN is a one-dimensional fully convolutional network, wherein each hidden layer maintains the same length as the input layer, achieved through zero-padding. This padding ensures that each successive layer remains the same length as the preceding one. To satisfy the second principle, TCN employs causal convolutions. A causal convolution is a specialized one-dimensional convolutional network where only elements from time $t$ and earlier are convolved to produce the output at time $t$. Fig. 15 demonstrates the structure of a causal convolutional network.

### Dilated Convolution:

TCN aims to effectively capture long-range dependencies in sequential data. A simple causal convolution can only consider a history that scales linearly with the depth of the network. This limitation would necessitate the use of large filters or an exceptionally deep network structure, which could hinder performance, particularly for tasks requiring a longer history.

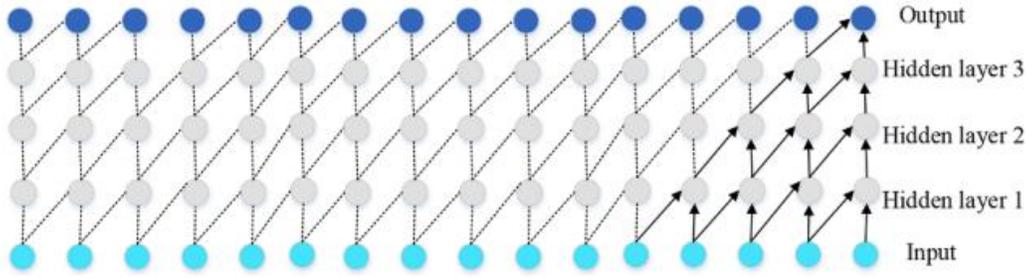

**Figure 15.** The structure of the causal convolutional network [87].

The depth of the network could lead to issues such as vanishing gradients, ultimately degrading network performance or causing it to plateau. To address these challenges, TCN employs dilated convolutions [90], which exponentially expand the receptive field, allowing the network to process large time series efficiently without a proportional increase in computational complexity. The architecture of a dilated convolutional network is depicted in Fig. 16.

By inserting gaps between the weights of the convolutional kernel, dilated convolutions effectively increase the network's receptive field while maintaining computational efficiency. The mathematical formulation of a dilated convolution is given by Eq. (11).

$$F(s) = (x *_d f)(s) = \sum_{i=0}^{k-1} f(i) \cdot x_{s-d \cdot i} \tag{11}$$

where $d$ is the dilation rate, $k$ is the size of the filter, and $s - d \cdot i$ accounts for the direction of the past. Dilation is the same as adding a fixed step in between each pair of neighboring filter taps. When $d = 1$, dilated convolution becomes a regular convolution. As $d$ increases, the output at the higher layers reflects a broader range of inputs, improving performance on long-range dependencies in time series.

**Residual Connections**:

To construct a more expressive TCN model, it is essential to use small filter sizes and stack multiple layers. However, stacking dilated and causal convolutional layers increases the depth of the network, potentially leading to problems such as gradient decay or vanishing gradients during training. To mitigate these issues, TCN incorporates residual connections into the output layer. Residual connections facilitate the flow of data across layers by adding a shortcut path, allowing the network to learn residual functions, which are modifications to the identity mapping, rather than learning a full transformation. This approach has been shown to be highly effective in very deep networks.

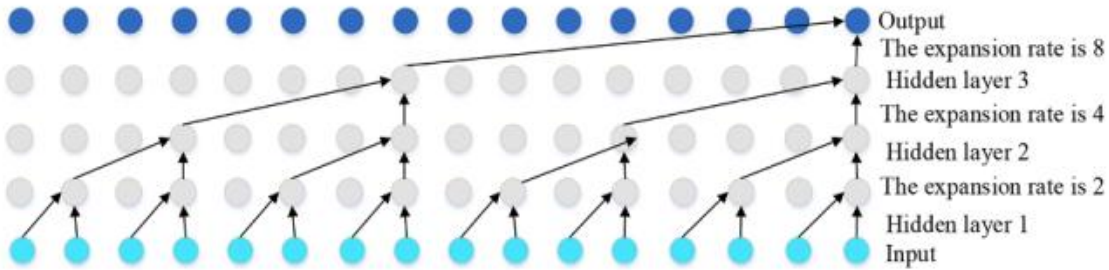

**Figure 16.** Dilated convolutional structure [87].

A residual block [46] has a branch that lead to a set of transformations F, whose outputs are appended to block's input x, as shown in Eq. (12).

$$o = Activation\left(x + F(x)\right) \tag{12}$$

This method enables the network to focus on learning residual functions rather than the entire mapping. The TCN residual block typically consists of two layers of dilated causal convolutions followed by a non-linear activation function, such as Rectified Linear Unit (ReLU). The convolutional filters within the TCN are normalized using weight normalization [91], and dropout [92] is applied to each dilated convolution layer for regularization, where an entire channel is zeroed out at each training step. In contrast to a conventional ResNet, where the input is directly added to the output of the residual function, TCN adjusts for differing input-output widths by performing an additional $1 \times 1$ convolution to ensure that the element-wise addition $\oplus$ operates on tensors of matching dimensions.

### 3.6 Kolmogorov-Arnold Network (KAN)

Kolmogorov-Arnold Networks (KANs) represent a promising alternative to traditional Multi-Layer Perceptrons (MLPs) by leveraging the Kolmogorov-Arnold theorem, a sophisticated mathematical framework that enhances the capacity of neural networks to process complex data structures. KANs were first introduced in 2024 by [93], with the goal of incorporating advanced mathematical theories into deep learning architectures to improve their performance on intricate tasks. While MLPs are inspired by the universal approximation theorem, KANs are motivated by the Kolmogorov-Arnold representation theorem [94], which states that any multivariate continuous function $f$ over a bounded domain can be expressed as a finite composition of simpler one-dimensional continuous functions:

$$f(x_1, \ldots, x_n) = \sum_{q=1}^{2n+1} \Phi_q\left(\sum_{p=1}^{n} \phi_{q,p}(x_p)\right) \tag{13}$$

where $\phi_{q,p}$ is a mapping $[0,1] \rightarrow \mathbb{R}$ and $\Phi_q$ is a mapping $\mathbb{R} \rightarrow \mathbb{R}$.

KAN maintain a fully connected structure like MLP, but with a key distinction: while MLP assign fixed activation functions to nodes (neurons), KAN assign learnable activation functions to edges (weights). Consequently, KAN do not employ traditional linear weight matrices; instead, each weight parameter is replaced by a learnable one-dimensional function parameterized as a spline. Unlike MLP, which apply non-linear activation functions at each node, KAN nodes only sum the incoming data, relying on the rich, learnable spline functions to introduce non-linearity. Although this approach might initially seem computationally expensive, KAN often result in significantly smaller computation graphs compared to MLP. Fig. 17 illustrates the structure of a KAN.

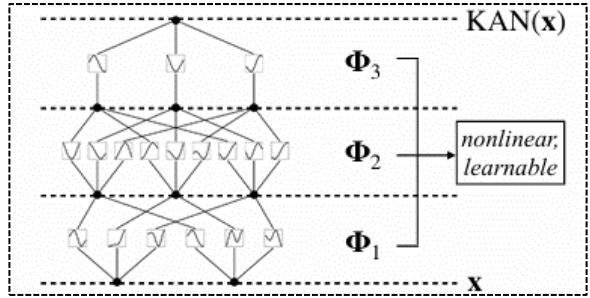

**Figure 17**. The structure of Kolmogorov-Arnold Network (KAN) [93].

The Kolmogorov-Arnold Network (KAN) can be expressed specifically as follows:

$$KAN(x) = (\Phi_{L-1} \circ \Phi_{L-2} \circ \cdots \circ \Phi_1 \circ \Phi_1)(x) \tag{14}$$

The transformation of each layer, $\Phi_l$, operates on the input $x_l$ to generate $x_{l+1}$, the input for the following layer, as follows:

$$x_{l+1} = \Phi_l(x_l) = \begin{pmatrix} \phi_{l,1,1}(\cdot) & \phi_{l,1,2}(\cdot) & \cdots & \phi_{l,1,n_l}(\cdot) \\ \phi_{l,2,1}(\cdot) & \phi_{l,2,2}(\cdot) & \cdots & \phi_{l,2,n_l}(\cdot) \\ \vdots & \vdots & \ddots & \vdots \\ \phi_{l,n_{l+1},1}(\cdot) & \phi_{l,n_{l+1},2}(\cdot) & \cdots & \phi_{l,n_{l+1},n_l}(\cdot) \end{pmatrix} x_l \tag{15}$$

Where each activation function $\phi_{l,j,i}$ is a spline, offering a rich, flexible response surface to inputs from the model:

$$spline(x) = \sum_i c_i B_i(x), \quad c_i \text{ are trainable coefficients} \tag{16}$$

Several variants of KANs have emerged to tackle specific challenges in various applications:

➤ **Convolutional KAN (CKAN) [95]:** CKAN is a pioneering alternative to standard CNN, which have significantly advanced the field of computer vision. Convolutional KAN integrate the non-linear activation functions of KAN into the convolutional layers, leading to a substantial reduction in the number of parameters and offering a novel approach to optimizing neural network architectures.

➤ **Temporal KAN (TKAN)** [96]: Temporal Kolmogorov-Arnold Networks combines the principles of KAN and Long Short-Term Memory (LSTM) networks to create an advanced architecture for time series analysis. Comprising layers of Recurrent Kolmogorov-Arnold Networks (RKANs) with embedded memory management, TKAN excels in multi-step time series forecasting. The TKAN architecture offers tremendous promise for improvement in domains needing one-step-ahead forecasting by solving the shortcomings of existing models in handling complicated sequential patterns [97, 98].

➤ **Multivariate Time Series KAN (MT-KAN)** [99]: MT-KAN is specifically designed to handle multivariate time series data. The primary objective of MT-KAN is to enhance forecasting accuracy by modeling the intricate interactions between multiple variables. MT-KAN utilizes spline-parametrized univariate functions to capture temporal relationships while incorporating methods to model cross-variable interactions.

➤ **Fractional KAN (fKAN) [100]:** fKAN is an enhancement of the KAN architecture that integrates the unique properties of fractional-orthogonal Jacobi functions into the network's basis functions. This method guarantees effective learning and improved accuracy by utilizing the special mathematical characteristics of fractional Jacobi functions, such as straightforward derivative equations, non-polynomial behavior, and activity for positive and negative input values.

➤ **Wavelet KAN (Wav-KAN) [101]:** The purpose of this innovative neural network design is to improve interpretability and performance by incorporating wavelet functions into the Kolmogorov-Arnold Networks (KAN) framework. Wav-KAN is an excellent way to capture complicated data patterns by utilizing wavelets' multiresolution analysis capabilities. It offers a reliable solution to the drawbacks of both recently suggested KANs and classic multilayer perceptrons (MLPs).

➤ **Graph KAN** [102]**:** This innovative model applies KAN principles to graph-structured data, replacing the MLP and activation functions typically used in Graph Neural Networks (GNNs) with KAN. This substitution enables more effective feature extraction from graph-like data structures.

## 4 Generative (Unsupervised) Deep Learning Models

Supervised machine learning is widely used in artificial intelligence (AI), while unsupervised learning remains an active area of research with numerous unresolved questions. However, recent advancements in deep learning and generative modeling have injected new possibilities into unsupervised learning. A rapidly evolving domain within computer vision research is generative models (GMs). These models leverage training data originating from an unknown data-generating distribution to produce novel samples that adhere to the same distribution. The ultimate goal of generative models is to generate data samples that closely resemble real data distribution [103].

Various generative models have been developed and applied in different contexts, such as Auto-Encoder [104], Generative Adversarial Network (GAN) [105], Restricted Boltzmann Machine (RBM) [106], and Deep Belief Network (DBN) [107].

### 4.1 Autoencoder

The concept of an autoencoder originated as a neural network designed to reconstruct its input data. Its fundamental objective is to learn a meaningful representation of the data in an unsupervised manner, which can have various applications, including clustering [104].

An autoencoder is a neural network that aims to replicate its input at its output. It consists of an internal hidden layer that defines a code representing the input data. The autoencoder network is comprised of two main components: an encoder function, denoted as $z = f(x)$, and a decoder function that generates a reconstruction, denoted as $r = g(z)$ [108]. The function $f(x)$ transforms a data point $x$ from the data space to the feature space, while the function $g(z)$ transforms $z$ from the feature space back to the data space to reconstruct the original data point $x$. In modern autoencoders, these functions $z = f(x)$ and $r = g(z)$ are considered as stochastic functions, represented as $p_{encoder}(z|x)$ and $p_{dencoder}(r|z)$, respectively, where $r$ denotes the reconstruction of $x$ [109]. Fig. 18 illustrates an autoencoder model.

Autoencoder models find utility in various unsupervised learning tasks, such as generative modeling [110], dimensionality reduction [111], feature extraction [112], anomaly or outlier detection [113], and denoising [114].

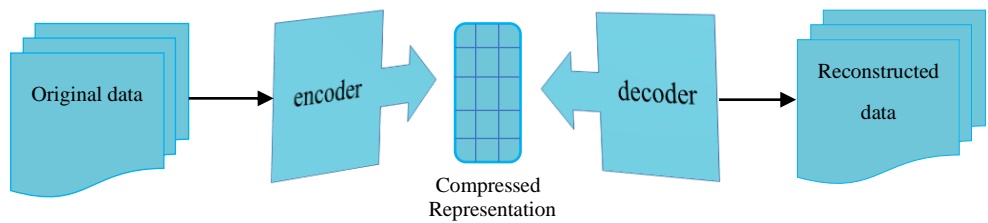

**Figure 18**. The structure of autoencoders.

In general, autoencoder models can be categorized into two major groups: Regularized Autoencoders, which are valuable for learning representations for subsequent classification tasks, and Variational Autoencoders [115], which can function as generative models. Examples of regularized autoencoder models include Sparse Autoencoder (SAE) [116], Contractive Autoencoder (CAE) [117], and Denoising Autoencoder (DAE) [118].

Variational Autoencoder (VAE) is a generative model that employs probabilistic distributions, such as the mean and variance of a Gaussian distribution, for data generation [104]. VAE provide a principled framework for learning deep latent-variable models and their associated inference models. The VAE consists of two coupled but independently parameterized models: the encoder or recognition model and the decoder or generative model. During "expectation maximization" learning iterations, the generative model receives an approximate posterior estimation of its latent random variables from the recognition model, which it uses to update its parameters. Conversely, the generative model acts as a scaffold for the recognition model, enabling it to learn meaningful representations of the data, such as potential class labels. In terms of Bayes' rule, the recognition model is roughly the inverse of the generative model [119].

### *4.2 Generative Adversarial Network (GAN)*

A notable neural network architecture for generative modeling, capable of producing realistic and novel samples on demand, is the Generative Adversarial Network (GAN), initially proposed by Ian Goodfellow in 2014 [105]. A GAN consists of two key components: a generative model and a discriminative model. The generative model aims to generate data that resemble real ones, while the discriminative model aims to differentiate between real and synthetic data. Both models are typically implemented using multilayer perceptrons [120]. Fig. 19 depicts the framework of a GAN, where a two-player adversarial game is played between a generator (G) and a discriminator (D). The generator's updating gradients are determined by the discriminator through an adaptive objective [121].

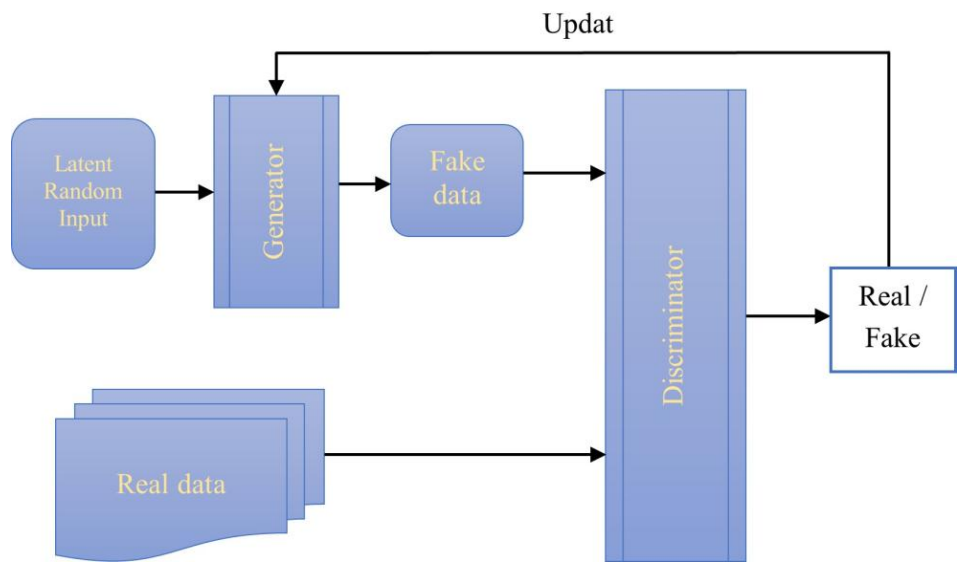

**Figure 19.** The framework of a GAN.

As previously mentioned, GANs operate based on principles derived from neural networks, utilizing a training set as input to generate new data that resembles the training set. In the case of GANs trained on image data, they can generate new images exhibiting human-like characteristics.

The following outlines the step-by-step operation of a GAN [122]:

1. The generator, created by a discriminative network, generates content based on the real data distribution.

2. The system undergoes training to increase the discriminator's ability to distinguish between synthesized and real candidates, allowing the generator to better fool the discriminator.

3. The discriminator initially trains using a dataset as the training data.

4. Training sample datasets are repeatedly presented until the desired accuracy is achieved.

5. The generator is trained to process random input and generate candidates that deceive the discriminator.

6.  Backpropagation is employed to update both the discriminator and the generator, with the former improving its ability to identify real images and the latter becoming more adept at producing realistic synthetic images.

7.  Convolutional Neural Networks (CNNs) are commonly used as discriminators, while deconvolutional neural networks are utilized as generative networks.

Generative Adversarial Networks (GANs) have introduced numerous applications across various domains, including image blending [123], 3D object generation [124], face aging [125], medicine [126, 127], steganography [128], image manipulation [129], text transfer [130], language and speech synthesis [131], traffic control [132], and video generation [133].

Furthermore, several models have been developed based on the Generative Adversarial Network (GAN) framework to address specific tasks. These models include Laplacian GAN (Lap-GAN) [134], Coupled GAN (Co-GAN) [120], Markovian GAN [135], Unrolled GAN [136], Wasserstein GAN (WGAN) [137], and Boundary Equilibrium GAN (BEGAN) [138], CycleGAN [139], DiscoGAN [140], Relativistic GAN [141], StyleGAN [142], Evolutionary GAN (E-GAN) [121], Bayesian Conditional GAN [143], Graph Embedding GAN (GE-GAN) [132].

### *4.3 Deep Belief Network (DBN)*

The Deep Belief Network (DBN) is a type of deep generative model utilized primarily in unsupervised learning to uncover patterns within large datasets. Consisting of multiple layers of hidden units, DBNs are adept at identifying intricate patterns and extracting features from data. Unlike discriminative models, DBNs exhibit a higher resistance to overfitting, making them well-suited for feature extraction from unlabeled data [144].

The stack of Restricted Boltzmann Machines (RBMs), which operate in an unsupervised learning framework, is a fundamental part of DBN. Every RBM in a DBN is made up of a hidden layer that contains latent representations and a visible layer that represents observable data features [145]. RBMs are trained layer by layer: first, each RBM is trained independently, and then all of the RBMs are fine-tuned together as a whole within the DBN.

During the forward pass, the activations represent the probability of an output given a weighted input. In the backward pass, the activations estimate the probability of inputs given the weighted outputs. Through iterative training of RBMs within a DBN, these processes converge to form joint probability distributions of activations and inputs, allowing the network to effectively capture the underlying data structure [146, 147]. Fig. 20 illustrates the schematic structure of a Deep Belief Network (DBN).

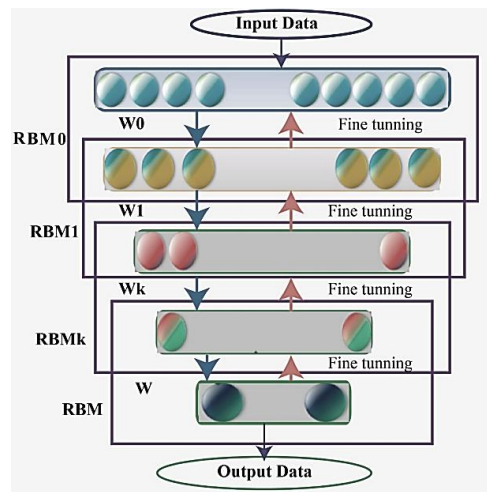

**Figure 20**. structure of a DBN model [145].

## 5 Transformer Architecture

The Transformer architecture was originally introduced by Vaswani et al. [148] in 2017 for machine translation and has since become a foundational model in deep learning, especially for natural language processing (NLP). The transformer functions as a self-attention encoder-decoder structure. The encoder consists of a stack of identical layers, and each layer consists of two sublayers. A multi-head self-attention mechanism is the first layer, while the other layer is a position-wise fully connected feed-forward network. Also, A normalizing layer [149] and residual connections [46] connect the multi-headed self-attention module's inputs and output. After that, a decoder uses the representation that the encoder produced to create an output sequence. A stack of identical layers makes up the decoder as well. The decoder adds a third sub-layer to each encoder layer in addition to the primary two, and this sub-layer handles multi-head attention over the encoder stack's output. Like the encoder, residual connections and a normalizing layer are used surrounding each of the sub-layers. The encoder and decoder's overall Transformer design is depicted in Fig. 21, left and right halves respectively [150, 151].

Traditional RNN-based Seq2Seq models could be replaced with attention layers. Using various projection matrices, the query, key, and value vectors in the self-attention layer are all produced from the same sequence [152]. RNN training takes a very long period because it is sequential and iterative. Transformer training, on the other hand, is parallel and enables all features to be learned concurrently, significantly improving computational efficiency and cutting down on the amount of time needed for model training [153].

**Multi-Head Attention:** In the Transformer model, a multi-headed self-attention mechanism is employed to enhance the model's ability to capture dependencies between elements in a sequence. The core principle of the attention mechanism is that every token in the sequence can aggregate information from other tokens, allowing the model to understand contextual relationships more effectively. This is achieved by mapping a query, a set of key-value pairs, and an output (each represented as vectors) to form an attention function. The output is computed as a weighted sum of the values, where the weights are determined by the compatibility function between the query and its corresponding key [148].

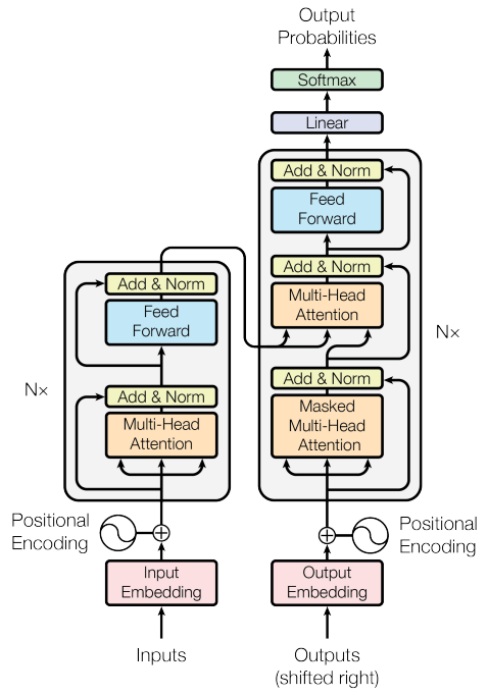

**Figure 21.** The architecture of the Transformer model [148].

Multi-head attention is equivalent to the blended of $n$ distinct scaled dot-product attention (self-attention). It can effectively process the three vectors Q, K, and V, in parallel to obtain the final result by combining and calculating. The formula is visible in Eq. (17).

$$\begin{cases} MultiHead\ (Q, K, V) = Concat(head_1, \dots, head_2)W^O \\ where\ head_i = Attention\ (QW_i^Q, KW_i^K, VW_i^V) \end{cases} \quad (17)$$

Where the projections are parameter matrices $W_i^Q \in \mathbb{R}^{d_{model} \times d_k}, W_i^K \in \mathbb{R}^{d_{model} \times d_k}, W_i^V \in \mathbb{R}^{d_{model} \times d_V}, and\ W^O \in \mathbb{R}^{hd_v \times d_{model}}.$

The main component of the transformer, scaled dot-product attention (self-attention), uses the weight of each sensor event in the input vector, which is represented by

$$Attention\ (Q, K, V) = softmax\left(\frac{QK^T}{\sqrt{d_k}}\right)V \quad (18)$$

The initial step in scaled dot-product attention is to convert the input data into an embedding vector and the three vectors of query vector (Q), key vector (K), and value vector (V) are then extracted from the embedding vectors. Next, a score is determined for every vector: score is equal to $Q \cdot K$. Score normalization (dividing by $\sqrt{d_k}$) is used for gradient stability. Next, the score is processed using the softmax activation function. The weighted score $v$ for every input vector is obtained by taking the softmax dot product value $v$. The final result is produced after summing. Scaled dot-product attention and multi-head attention are displayed in Fig. 22 [154].

**Position-wise Feed-Forward Networks:** Each encoder and decoder layer have a fully connected feed-forward network in addition to attention sub-layers. This feed-forward network is applied to each position independently and in the same way. This is made up of two linear transformations connected by a ReLU activation.

$$FFN(x) = max(0, xW_1 + b_1)W_2 + b_2 \quad (19)$$

**Positional Encoding:** Since the Transformer model does not rely on recurrence or convolution, it requires a way to capture the relative or absolute positions of tokens within a sequence to effectively utilize the sequence's order. To address this, positional encoding is introduced at the input level of both the encoder and decoder stacks. These positional encodings are added to the input embeddings, as they share the same dimensionality, $d_{model}$. This combination enables the model to incorporate positional information, allowing it to better understand the sequential nature of the data [148].

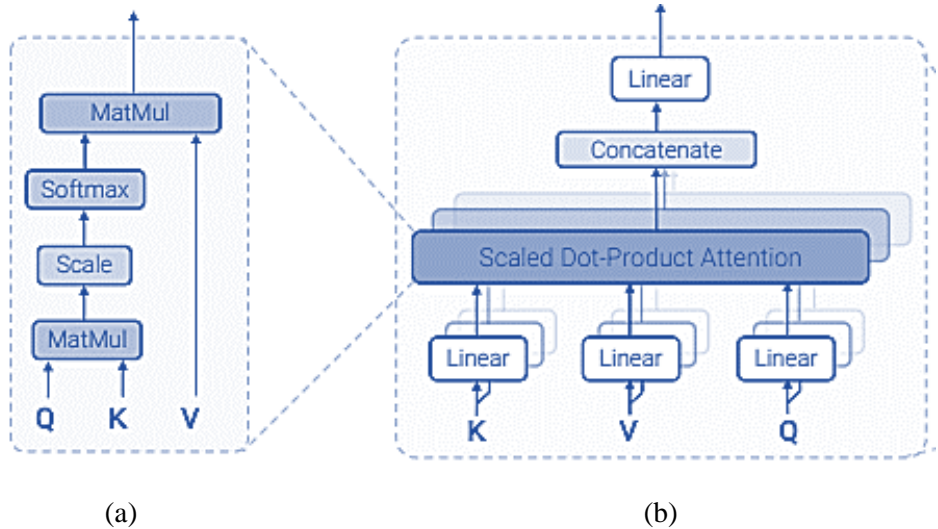

(a)                                          (b)

**Figure 22.** (a) Scaled Dot-Product Attention, (b) Multi-Head Attention.

Positional encodings in transformer architecture were achieved by using sine and cosine functions of various frequencies:

$$\begin{cases} PE_{(pos,2i)} = sin(pos/10000^{2i/d_{model}}) \\ PE_{(pos,2i+1)} = cos(pos/10000^{2i/d_{model}}) \end{cases} \tag{20}$$

where $pos$ is the position and $i$ is the dimension. Every dimension of the positional encoding has a sinusoidal relationship. The wavelengths range from $2\pi\ to\ 10000 \cdot 2\pi$ in a geometric development. This function was selected because it would make it simple for the model to learn how to attend to relative positions, since for any fixed offset $k$, $PE_{pos+k}$ can be expressed as a linear function of $PE_{pos}$.

### 5.1 Transformer Variants

The Transformer architecture has proven to be highly versatile, with numerous variants developed to address specific challenges across different domains. Typically, Transformers are pre-trained on large datasets using unsupervised methods to learn general representations, which are then fine-tuned on specific tasks using supervised learning. This hybrid approach leverages the strengths of both learning paradigms. Some notable Transformer variants include:

➤ **Bidirectional Encoder Representations from Transformers (BERT)** [155]: A multi-layer bidirectional Transformer encoder for unsupervised pre-training in natural language understanding (NLU) tasks.

➤ **Generative pre-training Transformer (GPT)** [156, 157]: A type of Transformer model developed by OpenAI that excels in natural language processing (NLP) tasks through unsupervised pre-training followed by supervised fine-tuning.

➤ **Transformer-XL** [158]: It is proposed for language modeling to permit learning reliance beyond a set length without compromising temporal coherence. Transformer-XL (Transformer-Extra Long) comprises a unique relative positional encoding method and a segment-level recurrence mechanism. This approach not only makes it possible to record longer-term dependencies, but also fixes the issue of context fragmentation.

➤ **XLNet** [159]: It is a generalized autoregressive (AR) pretraining technique that combines the benefits of autoencoding (AE) and autoregressive (AR) techniques with a permutation language modeling aim. XLNet's neural architecture, which integrates Transformer-XL and the two-stream attention mechanism, is built to function effortlessly with the autoregressive (AR) objective.

➤ **Fast Transformer** [160]: It introduces multi-query attention as an alternative to multi-head attention. This approach reduces memory bandwidth requirements, leading to increased processing speed.

➤ **Multimodal Transformer (MulT)** [161]: It is designed for analyzing human multimodal language. At the heart of MulT is the crossmodal attention mechanism, which provides a latent crossmodal adaptation that fuses multimodal information by directly attending to low-level features in other modalities.

➤ **Vision Transformer (ViT)** [162]: An innovative approach based on Transformer structure for visual tasks like image classification.

➤ **Pyramid Vision Transformer (PVT)** [163]: An Transformer framework for complex prediction tasks like semantic segmentation and object recognition.

➤ **Swin Transformer** [164]: A hierarchical transformer that uses shifted windows to construct its representation. A wide variety of vision tasks, including semantic segmentation, object detection, and image classification, may be performed with Swin Transformer.

➤ **Tokens-to-Token Vision Transformer (T2T-ViT)** [165]: A vision transformer that can be trained from scratch on ImageNet. T2T-ViT overcomes ViT's drawbacks by accurately modeling the structural information of images and enhancing feature richness.

- ➤ **Transformer in Transformer (TNT)** [166]: A vision transformer for visual recognition. Both local and global representations are extracted by the TNT architecture through the use of an inner transformer and an outer transformer.
- ➤ **PyramidTNT** [167]: A improved TNT model which used pyramid architecture, and convolutional stem in order to greatly enhance the original TNT model.
- ➤ **Switch Transformers** [168]: It is suggested as a straightforward and computationally effective method of increasing a Transformer model's parameter count.
- ➤ **ConvNeXt** [169]: A redesigned transformer architecture that makes use of the transformer attention mechanism and incorporates convolutional layers into the encoder and decoder modules to extract spatially localized data.
- ➤ **Evolutionary Algorithm Transformer (EATFormer)** [170]: An improved vision transformer influenced by an evolutionary algorithm.

## 6 Deep Reinforcement Learning

Reinforcement learning (RL) is a machine learning approach that deals with sequential decision-making, aiming to map situations to actions in a way that maximizes the associated reward. Unlike supervised learning, where explicit instructions are given after each system action, in the RL framework, the learner, known as an agent, is not provided with explicit guidance on which actions to take at each timestep $t$. The RL agent must explore through trial and error to determine which actions yield the highest rewards [171]. Furthermore, unlike supervised learning, where the correct output is obtained and the model is updated based on the loss or error, RL uses gradients without a differentiable loss function to teach a model to explore randomly and learn to make optimal decisions [172]. Fig. 23 depicts the agent-environment interaction in reinforcement learning (RL). The standard theoretical framework for RL is based on a Markov Decision Process (MDP), which extends the concept of a Markov process and is used to model decision-making based on states, actions, and rewards [173].

Deep reinforcement learning combines the decision-making capabilities of reinforcement learning with the perception function of deep learning. It is considered a form of "real AI" as it aligns more closely with human thinking. Fig. 24 illustrates the basic structure of deep reinforcement learning, where deep learning processes sensory inputs from the environment and provides the current state data. The reinforcement learning process then links the current state to the appropriate action and evaluates values based on anticipated rewards [174, 175].

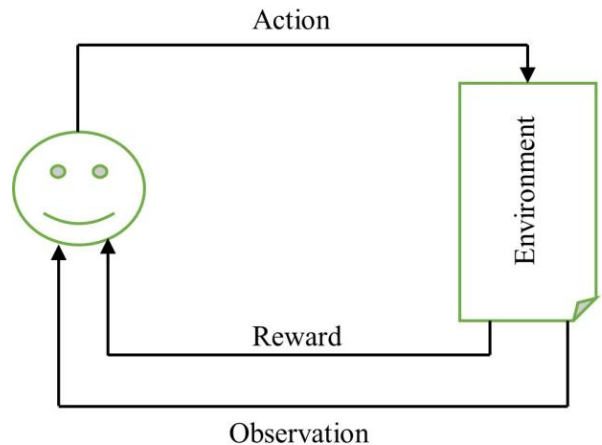

**Figure 23**.    Agent-Environment interaction in RL.

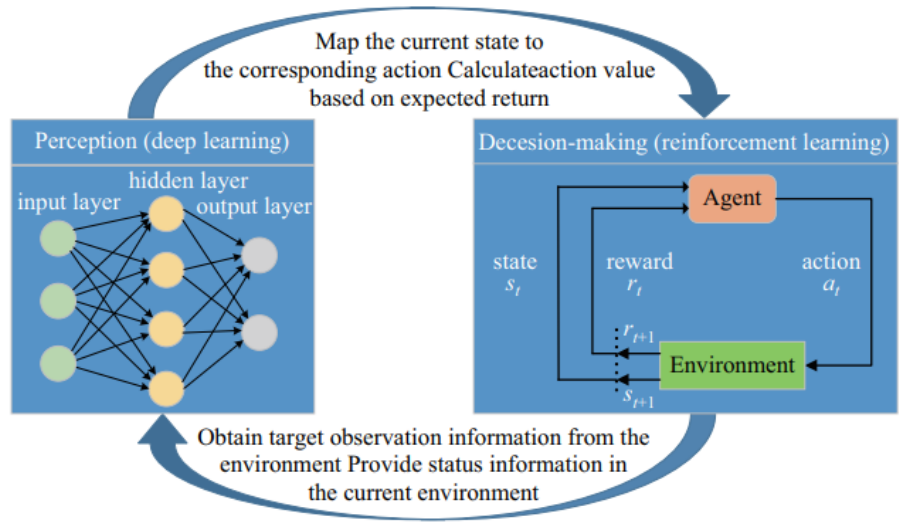

**Figure 24**.    Basic structure of Deep Reinforcement Learning (DRL) [174].

One of the most renowned deep reinforcement learning models is the Deep Q-learning Network (DQN) [176], which directly learns policies from high-dimensional inputs using Convolutional Neural Network (CNN). Other common models in deep reinforcement learning include Double DQN [177], Dueling DQN [178], and Monte Carlo Tree Search (MCTS) [179].

Deep reinforcement learning (DRL) models find applications in various domains, such as video game playing [180, 181], robotic manipulation [182, 183], image segmentation [184, 185], video analysis [186, 187], energy management [188, 189], and more.

## 7 Deep Transfer Learning

Deep neural networks have significantly improved performance across various machine learning tasks and applications. However, achieving these remarkable performance gains often requires large amounts of labeled data for supervised learning, as it relies on capturing the latent patterns within the data [190]. Unfortunately, in certain specialized domains, the availability of sufficient training data is a major challenge. Constructing a large-scale, high-quality annotated dataset is costly and time-consuming [191].

To address the issue of limited training data, transfer learning (TL) has emerged as a crucial tool in machine learning. The concept of transfer learning finds its roots in educational psychology, where the theory of generalization suggests that transferring knowledge from one context to another is facilitated by generalizing experiences. To achieve successful transfer, there needs to be a connection between the two learning tasks. For example, someone who has learned to play the violin is likely to learn the piano more quickly due to the shared characteristics between musical instruments [192]. Fig. 25 depicts the learning process of transfer learning. Deep transfer learning (DTL) makes use of the learning experience to reduce the time and effort needed to train large networks as well as the time and effort needed to create the weights for an entire network from scratch [193].

With the growing popularity of deep neural networks in various fields, numerous deep transfer learning techniques have been proposed. Deep transfer learning can be categorized into four main types based on the techniques employed [191]: instances-based deep transfer learning, mapping-based (feature-based) deep transfer learning, network-based (model-based) deep transfer learning, and adversarial-based deep transfer learning.

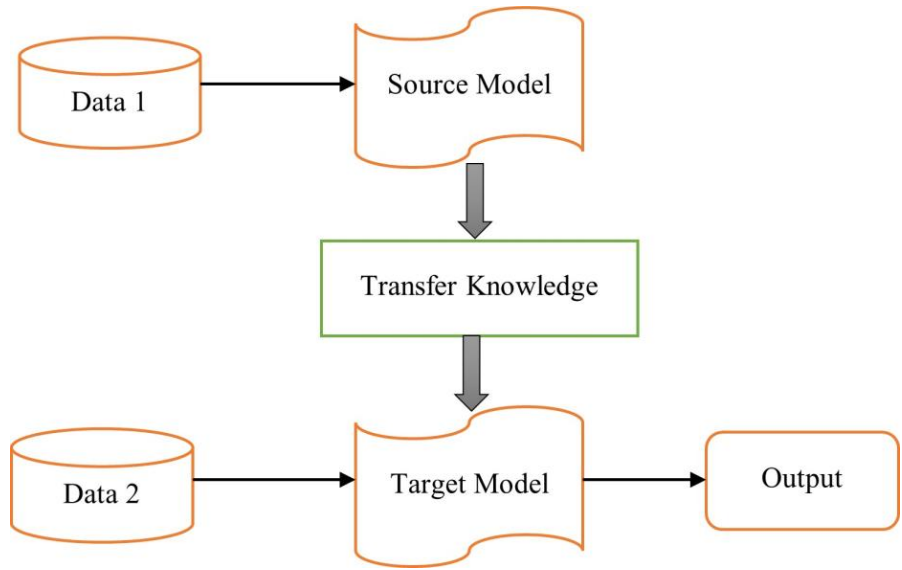

**Figure 25.** Learning process of transfer learning.

Instances-based deep transfer learning involves selecting a subset of instances from the source domain and assigning appropriate weight values to these selected instances to supplement the training set in the target domain. Algorithms such as TaskTrAdaBoost [194] and TrAdaBoost.R2 [195] are well-known approaches based on this strategy.

Mapping-based deep transfer learning focuses on mapping instances from both the source and target domains into a new data space, where instances from the two domains exhibit similarity and are suitable for training a unified deep neural network. Successful methods based on this approach include Extend MMD (Maximum Mean Discrepancy) [196], and MK-MMD (Multiple Kernel variant of MMD) [197].

Network-based (model-based) deep transfer learning involves reusing a segment of a pre-trained network from the source domain, including its architecture and connection parameters, and applying it to a deep neural network in the target domain. These model-based approaches are highly effective for domain adaptation between source and target data by adjusting the network (model), making them the most widely adopted strategies in deep transfer learning (DTL). Remarkably, these methods can even adapt target data that is significantly different from the source data [198].

Network-based (model-based) approaches in deep transfer learning typically involve pre-training, freezing, fine-tuning, and adding new layers. Pre-trained models consist of layers from a deep learning network (DL model) that have been trained using source data. Two key methods for training a model with target data are freezing and fine-tuning. These methods involve using some or all layers of a pre-defined model. When layers are frozen, they retain fixed parameters/weights from the pre-trained model. In contrast, fine-tuning involves initializing parameters and weights with pre-trained values instead of starting with random values, either for the entire network or specific layers [198].

A recent advancement in model-based deep transfer learning is Progressive Neural Networks (PNNs). This strategy involves the freezing of a pre-trained model and integrating new layers specifically for training on target data [199]. The concept behind progressive learning is grounded in the idea that acquiring a new skill necessitates leveraging existing knowledge. This mirrors the way humans learn new abilities. For instance, a child learns to run by employing all the skills acquired during crawling and walking. PNN constructs a new model for each task it encounters.

Each freshly generated model is interconnected with all others, aiming to learn a new task by applying the knowledge accumulated from preceding models.

Adversarial-based methods focus on gathering transferable features from both the source and target data by leveraging logical relationships or rules acquired in the source domain. Alternatively, they may utilize techniques inspired by generative adversarial networks (GANs) [200].

These deep transfer learning techniques have proven to be effective in overcoming the challenge of limited training data, enabling knowledge transfer across domains, and facilitating improved performance in various applications such as image classification [201, 202], speech recognition [203, 204], video analysis [205, 206], signal processing [207, 208], and other.

In transfer learning, several popular pre-trained deep learning models are frequently used, including Xception [52], MobileNet [53], DenseNet [55], EfficientNet [57], NasNet [209], and among others. These models are initially trained on large-scale datasets like ImageNet, and their learned weights are then transferred to a target domain. The architectures of these networks reflect a broader trend in deep learning design, transitioning from manually crafted by human experts to automatically optimized patterns. This evolution focuses on striking a balance between model accuracy and computational complexity [210].

## 8 Hybrid Deep Learning Models

Hybrid deep learning architectures, which integrate elements from various deep learning models, demonstrate significant potential in enhancing performance. By combining different fundamental generative or discriminative models, the following three categories of hybrid deep learning models can be particularly effective for addressing real-world problems:

- Combination of various supervised models to extract more relevant and robust features, such as CNN+LSTM or CNN+GRU. By leveraging the strengths of different architectures, these hybrid models effectively capture both spatial and temporal dependencies within the data.

- Integrating various types of generative models, such as combining Autoencoders (AE) with Generative Adversarial Networks (GANs), to harness their strengths and enhance performance across a range of tasks.

- Integrating the capabilities of generative models with supervised models to leverage the strengths of both approaches can significantly enhance performance on various tasks. This hybrid strategy improves feature learning, data augmentation, and model robustness. Examples of such combinations include DBN+MLP, GAN+CNN, AE+CNN, and so on.

## 9 Application of Deep Learning

In recent years, deep learning has demonstrated remarkable effectiveness across a wide range of applications, tackling various challenges in fields including healthcare, computer vision, speech recognition, natural language processing (NLP), e-learning, smart environments, and more. Fig. 26 highlights several potential real-world application areas of deep learning.

Five useful categories have been established for these applications: classification, detection, localization, segmentation, and regression [10]. A concept called classification divides a collection of facts into classes. Detection typically involves recognizing objects and their boundaries within images, videos, or other data types. Localization refers to the process of identifying and determining the position of specific objects or features within an image or other types of data. Segmentation involves dividing an image or dataset into distinct regions or segments, with each segment representing a particular object or feature of interest. Regression is used to model and analyze the relationships between a dependent variable and one or more independent variables. It predicts continuous outcomes based on input features.

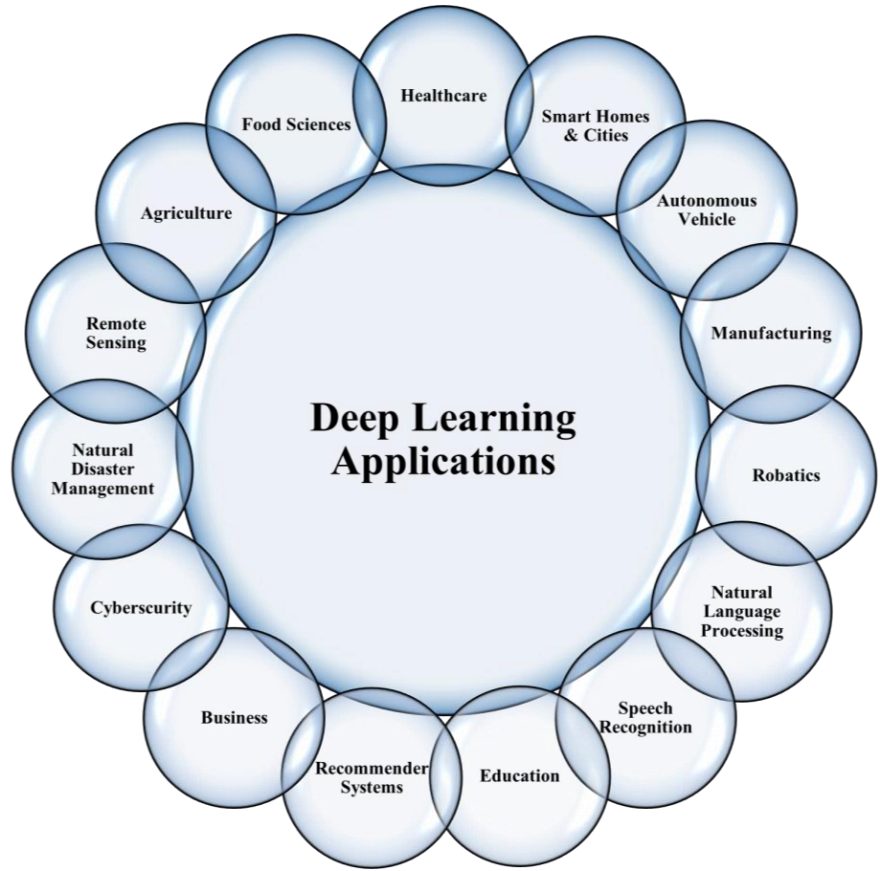

**Figure 26.** Numerous possible domains for deep learning applications in the real world.

However, each real-world application area has its own specific goals and requires particular tasks and deep learning techniques. Table 1 provides a summary of various deep learning tasks and methods applied across multiple real-world application domains.

**Table 1**: A summary of the practical applications of deep learning models in real-world domains.

| Application Setting | Tasks | Models | Reference |
|---|---|---|---|
| **Smart Homes & Smart Cities** | Human Activity Recognition | CNN+LSTM | [211] |
| | Smart Energy Management | Reinforcement learning | [212] |
| | Traffic Management | GRU based | [213] |
| | Waste Management | CNN based | [214] |
| | Smart Parking System | Stacked GRU+LSTM | [215] |
| **Education** | Student Engagement Detection | DenseNet self-attention | [216] |
| | Student Affective States Recognition | ConvNeXt + GRU | [82] |
| | Automatic Attendance System | CNN+LSTM | [217] |
| | Automated Exam Control | CNN based (VGG) | [218] |
| **Healthcare** | Medical Image Analysis | Vision transformer | [219] |
| | Early Disease Detection | InceptionV3 | [220] |

|  | Remote Patient Monitoring | CNN based | [221] |
|---|---|---|---|
|  | Analyze Genomic Data | Transfer learning based | [222] |
| **Natural Language Processing (NLP)** | Question Answering Systems | BERT based | [223] |
|  | Sentiment Analysis | Transformer based | [224] |
|  | Text Summarization | Attentional LSTM | [225] |
| **Speech Recognition** | Speech Emotion Recognition | LSTM+CNN | [226] |
|  | Automatic Speech Translation | Deep transfer learning | [203] |
| **Agriculture** | Plant Disease Detection | ViT +CNN | [227] |
|  | Precision Agriculture | GRU+CNN | [228] |
|  | Smart Irrigation System | Autoencoders, GAN | [229] |
|  | Soil Quality Prediction | CNN | [230] |
| **Natural Disaster Management** | Earthquake Prediction | CNN+RNN | [231] |
|  | Flood Forecasting | Attention GRU | [232] |
|  | Tsunami Prediction | LSTM based | [233] |
| **Remote Sensing** | Land Cover Classification | Extended ViT | [234] |
|  | Investigation Wildfire Area | CNN based | [235] |
|  | Deforestation Detection | Transformer based | [236] |
| **Cybersecurity** | Intrusion Detection | CNN+ Bi-LSTM | [237] |
|  | Malware Detection | LSTM based | [238] |
|  | Phishing Detection | LSTM+CNN | [239] |
|  | Credit Card Fraud Detection | Deep Autoencoder | [240] |
|  | Biometric Authentication | CNN+LSTM | [241] |
| **Recommender Systems** | Context-Aware Recommendation | RNN based | [242] |
|  | Sequential Recommendation | LSTM based | [243] |
|  | Multimodal Recommendation | CNN based | [244] |
| **Business** | Purchase Behavior Prediction | RNN based | [245] |
|  | Loan Default Prediction | CNN based | [246] |
|  | Stock Trend Prediction | Bi-LSTM | [247] |
| **Autonomous Vehicles** | Object Detection | Swin transformer +CNN | [248] |
|  | Pedestrian Detection | Deep CNN | [249] |
|  | Localization And Mapping | CNN-GRU | [250] |
|  | Lane Detection & Path Planning | CNN based | [251] |
| **Manufacturing** | Defect Detection | Transformer based | [252] |
|  | Predictive Maintenance | LSTM, GRU, CNN | [253] |
|  | Process Optimization | Reinforcement learning | [254] |
|  | Supply Chain Optimization | LSTM | [255] |
| **Robotics** | Robotic Grasping | Reinforcement learning | [256] |
|  | Tracking And Motion Planning | Reinforcement learning | [257] |
|  | Human-Robot Interaction | RNN based | [258] |

## 10 Deep Learning Challenges

While deep learning models have achieved remarkable success across various domains, they also come with significant challenges. Below are some of the most critical challenges, followed by potential solutions to address them.

### 10.1 Insufficient Data

Deep learning models require large amounts of data to perform well. The performance of these models typically improves as the volume of data increases. However, in many cases, sufficient data may not be available, making it difficult to train deep learning models effectively [10].

Three possible approaches may be used to appropriately handle the insufficient data problem. The first method is Transfer Learning (TL), which is used to DL models by reusing pre-trained model pieces in new models. We thoroughly reviewed the transfer learning strategy in section 7.

Data augmentation is the second method of gathering additional data. The goal of data augmentation is to improve the trained models' capacity for generalization. Generalization is necessary for networks to overcome small datasets or datasets with unequal class distributions, and it is especially crucial for real-world data [259]. There are several strategies for augmenting data, and each one is contingent upon the characteristics of the datasets [260]. A few of these techniques are geometric transformations [261], Mixup augmentation [262], Random oversampling [263], Feature space augmentation [264], generative data augmentation [265], and many more.

The third approach considers using simulated data to increase the training set's volume. If you have a good understanding of the physical process, you can sometimes build simulators from it. Consequently, the outcome will include simulating as much data as necessary [10, 266].

### 10.2 Imbalanced Data

In real-world situations, particularly in those that deep learning models address, the issue of class imbalance is common. If the majority of instances in the data set belong to one class and the remaining instances belong to the other class, then there is a class imbalance in a binary classification scenario. In multi-class, multi-label, multi-instance learning as well as in regression difficulties and other situations, class imbalances are present and are actually reinforced [267].

It has been determined that there are three main approaches to addressing imbalanced data: data-level techniques, algorithm-level techniques, and hybrid techniques. The focus of data-level techniques is to add or remove samples from training sets in order to balance the data distributions. These techniques balance the data distributions by adding new samples to the minority class (oversampling) or removing samples from the majority class (undersampling) [268, 269]. A variety of oversampling techniques, including Synthetic Minority Over-sampling Technique (SMOTE) [270], Borderline-SMOTE [271], Adaptive Synthetic (ADASYN) [272], SVM (Support Vector Machine)-SMOTE [273], Majority Weighted Minority Oversampling Technique (MWMOTE) [274], Sampling With the Majority (SWIM) [275], Reverse-SMOTE (R-SMOTE) [276], Constrained Oversampling (CO) [277], SMOTE Based on Furthest Neighbor Algorithm (SOMTEFUNA) [278], and many more can be used to solve imbalanced data problems. Also, there are several techniques for undersampling, including EasyEnsemble [279], BalanceCascade [279], Inverse Random Undersampling [280], MLP-based Undersampling Technique (MLPUS) [281], and others.

Algorithm-level approaches modify existing learning algorithms to mitigate the bias towards the majority class. These techniques require specialized knowledge of both the application domain and the learning algorithm to diagnose why a classifier fails under imbalanced class distributions [268]. Two of the most commonly used methods in this context are Cost-Sensitive Learning [282, 283] and One-Class Learning [284].

The third approach consists of hybrid methods, which combine algorithm-level techniques with data-level methods in the appropriate way. Hybridization is required to address issues with algorithm and data-level approaches and improve classification accuracy [285].

### 10.3 Overfitting

Overfitting occurs when a deep learning model learns the systematic and noise components of the training data to the point that it adversely affects the model's performance on new data. In fact, overfitting occurs as a result of noise, the small size of the training set, and the complexity of the classifiers. Overfitted models tend to memorize all the data, including the inevitable noise in the training set, rather than understanding the underlying patterns in the data [24]. Overfitting is addressed with methods including dropout [92], weight decay [286], batch normalization [287, 288], regularization [289], data augmentation, and others, although determining the ideal balance is still difficult.

### 10.4 Vanishing and Exploding Gradient

In deep neural networks, the computation of gradients is propagated layer by layer, leading to a phenomenon known as the vanishing or exploding gradient problem. As gradients are backpropagated through the network, they can exponentially diminish or grow, respectively, causing significant issues in training. When gradients vanish, the weights of the network are adjusted so minimally that the model's learning process becomes exceedingly slow, potentially stalling altogether. Conversely, exploding gradients can cause weights to be updated excessively, leading to instability and divergence during training. This problem is particularly pronounced with non-linear activation functions such as sigmoid and tanh, which compress the output into a narrow range, further exacerbating the issue by limiting the gradient's magnitude. Consequently, the model struggles to learn effectively, especially in deep networks where gradients must pass through many layers [8].

To mitigate the vanishing and exploding gradient problem, several strategies have been developed. One effective approach is to use the Rectified Linear Unit (ReLU) activation function, which does not saturate and therefore helps to maintain the gradient flow throughout the network [290]. Proper weight initialization techniques, such as Xavier initialization [291] can also reduce the likelihood of gradient issues by ensuring that initial weights are set in a way that prevents gradients from becoming too small or too large [292]. Another solution is batch normalization, which normalizes the inputs of each layer to maintain a stable distribution of activations throughout training. By doing so, batch normalization helps to alleviate the vanishing gradient problem and can accelerate convergence by reducing internal covariate shifts. Overall, addressing the vanishing and exploding gradient problem is crucial for training deep neural networks effectively, enabling them to learn complex patterns without succumbing to instability or inefficiency [288].

### 10.5 Catastrophic Forgetting

Catastrophic forgetting is a critical challenge in the pursuit of artificial general intelligence within neural networks. It occurs when a model, after being trained on a new task, loses its ability to perform previously learned tasks. This phenomenon is particularly problematic in scenarios where a model is expected to learn sequentially across multiple tasks without forgetting earlier ones, such as in lifelong learning or continual learning applications. The root cause of catastrophic forgetting lies like neural networks, which update their weights based on new training data. When trained on a new task, the model adjusts its parameters to optimize performance on that task, often at the expense of previously acquired knowledge. As a result, the model may exhibit excellent performance on the most recent task but perform poorly on earlier ones, effectively "forgetting"

them [293].

Several strategies have been proposed to address catastrophic forgetting. One such approach is Elastic Weight Consolidation (EWC) [294], which penalizes changes to the weights that are important for previous tasks, thereby preserving learned knowledge while allowing the model to adapt to new tasks. Incremental Moment Matching (IMM) ) [295] is another technique that merges models trained on different tasks into a single model, balancing the performance across all tasks. The iCaRL (incremental Classifier and Representation Learning) [296] method combines classification with representation learning, enabling the model to learn new classes without forgetting previously learned ones. Additionally, the Hard Attention to the Task (HAT) [293] approach employs task-specific masks that prevent interference between tasks, reducing the likelihood of forgetting.

### *10.6 Underspecifcation*

Underspecification is an emerging challenge in the deployment of machine learning (ML) models, particularly deep learning (DL) models, in real-world applications. It refers to the phenomenon where an ML pipeline can produce a multitude of models that all perform well on the validation set but exhibit unpredictable behavior in deployment. This issue arises because the pipeline's design does not fully specify which model characteristics are critical for generalization in real-world scenarios. The underspecification problem is often linked to the high degrees of freedom inherent in ML pipelines. Factors such as random seed initialization, hyperparameter selection, and the stochastic nature of training can lead to the creation of models with similar validation performance but divergent behaviors in production. These differences can manifest as inconsistent predictions when the model is exposed to new data or deployed in environments different from the training conditions [297].

Addressing underspecification requires rigorous testing and validation beyond standard metrics. Stress tests, as proposed by D'Amour et al. [297], are designed to evaluate a model's robustness under various real-world conditions, identifying potential failure points that may not be apparent during standard validation. These tests simulate different deployment scenarios, such as varying input distributions or environmental changes, to assess how the model's predictions might vary. Moreover, some researches have been conducted to analyze and mitigate underspecification across different ML tasks [298, 299].

## 11 Analysis of Deep Learning Models

This section details the methodology used in this study, which focuses on applying and evaluating various deep learning models for classification tasks across three distinct datasets. For our experimental analysis, we utilized three publicly available datasets: IMDB [300], ARAS [301], and Fruit-360 [302]. The objective is to conduct a comparative analysis of the performance of these deep learning models.

The IMDB dataset, which stands for Internet Movie Database, provides a collection of movie reviews categorized as positive or negative sentiments. ARAS is a dataset comprising annotated sensor events for human activity recognition tasks. Fruit-360 is a dataset consisting of images of various fruit types for classification purposes.

We began by evaluating eight different models: CNN, RNN, LSTM, Bidirectional LSTM, GRU, Bidirectional GRU, TCN, and Transformer on the IMDB and ARAS datasets. Our analysis aimed to compare the performance of these deep learning models across diverse datasets. The CNN model (Convolutional Neural Network) is particularly effective in capturing spatial dependencies, making it suitable for tasks involving structured data. RNN (Recurrent Neural Network) is well-suited for sequential data analysis, while LSTM (Long Short-Term Memory) and GRU (Gated Recurrent Unit) models are designed to capture long-term dependencies in sequential data. The Bidirectional LSTM and Bidirectional GRU models provide an additional advantage by processing information in both forward and backward directions.

Additionally, we evaluated eight different CNN-based models: VGG, Inception, ResNet, InceptionResNet, Xception, MobileNet, DenseNet, and NASNet for the classification of fruit images using the Fruit-360 dataset. Given that image data is not sequential or time-dependent, recurrent models were not suitable for this task. CNN-based models are particularly effective for image analysis because of their ability to capture spatial dependencies. Moreover, the faster training time of CNN models is due to their parallel processing capabilities, which allow for efficient computation on GPU (Graphics Processing Unit), thereby accelerating the training process.

To evaluate the performance of these models, we employed assessment metrics such as accuracy, precision, recall, and F1-measure. Accuracy measures the overall correctness of the model's predictions, while precision evaluates the proportion of correctly predicted positive instances. Recall assesses the model's ability to correctly identify positive instances, and F1-measure provides a balanced measure of precision and recall.

$$Accuracy = \frac{Tp + Tn}{Tp + Tn + Fp + Fn} \tag{21}$$

$$Precision = \frac{Tp}{Tp + Fp} \tag{22}$$

$$Recall = \frac{Tp}{Tp + Fn} \tag{23}$$

$$F1 - Score = 2 \times \frac{Recall \times Precision}{Recall + Precision} \tag{24}$$

Where $Tp$ = True Positive, $Tn$ = True Negative, $Fp$ = False Positive, and $Fn$ = False Negative.

By conducting a comprehensive analysis using these metrics, we can gain insights into the strengths and weaknesses of each deep learning model. This comparative evaluation enables us to identify the most effective model for specific datasets and applications, ultimately advancing the field of deep learning and its practical applications.

All experiments were conducted on a GeForce RTX 3050 GPU (Graphics Processing Unit) with 4 Gigabyte of RAM (Random Access Memory).

### 11.1 Methodology and Experiments on IMDB Dataset

The IMDB dataset is a widely used dataset for sentiment analysis tasks. It consists of movie reviews along with their corresponding binary sentiment polarity labels. The dataset contains a total of 50,000 reviews, evenly split into 25,000 training samples and 25,000 testing samples. There is an equal distribution of positive and negative labels, with 25,000 instances of each sentiment. To reduce the correlation between reviews for a given movie, only 30 reviews are included in the dataset [300]. Positive reviews often contain words like "great," "well," and "love," while negative reviews frequently use words like "bad" and "can't." However, certain words such as "one," "character," and "well" appear frequently in both positive and negative reviews, although their usage may differ in terms of frequency between the two sentiment classes [72].

In our analysis, we employed eight different deep learning models including CNN, RNN, LSTM, Bidirectional LSTM, GRU, Bidirectional GRU, TCN, and Transformer for sentiment classification using the IMDB dataset. Fig. 27 presents a structural overview of the deep learning model intended for analyzing the performance of eight different models on the IMDB dataset.

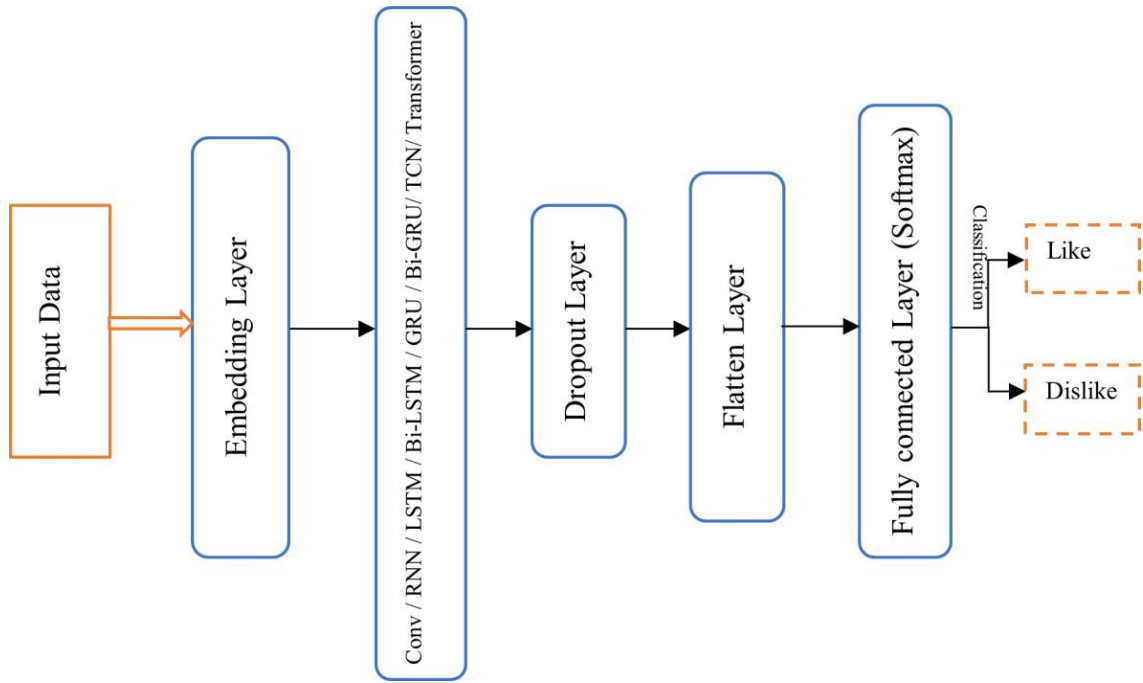

**Figure 27.**  The structural for analysis of different deep learning models on IMDB dataset

In this architecture, text data is first passed through an embedding layer, which transforms the high-dimensional, sparse input into dense, lower-dimensional vectors of real numbers. This allows the model to capture semantic relationships within the data. In the second layer, one of eight models: CNN, RNN, LSTM, Bi-LSTM, GRU, Bi-GRU, TCN, or Transformer is employed for feature extraction and data training. This layer is crucial for capturing patterns and dependencies in the data. Following this, a dropout layer is included to address the issue of overfitting by randomly deactivating a portion of the neurons during training, which helps improve the model's generalization. Subsequently, the multi-dimensional vector turns into a one-dimensional vector using a flatten layer, enabling it to work with fully connected layers. Finally, the output is passed through a fully connected (Dense) layer, which uses a Softmax function for classification, converting the model's predictions into probabilities for each class.

Building a neural network with high accuracy necessitates careful attention to hyperparameter selection, as these adjustments significantly influence the network's performance. For example, setting the number of training iterations too high can lead to overfitting, where the model performs well on the training data but poorly on unseen data. Another critical hyperparameter is the learning rate, which affects the rate of convergence during training. If the learning rate is too high, the network may converge too quickly, potentially overshooting the global minimum of the loss function. Conversely, if the learning rate is too low, the convergence process may become excessively slow, prolonging training. Therefore, finding the optimal balance of hyperparameters is essential for maximizing the network's performance and ensuring effective learning.

In the experiment phase, consistent parameters were applied across all models to ensure a standardized comparison. The parameters were set as follows: epochs = 30, batch size = 64, dropout = 0.2, with the loss function set to "Binary Crossentropy," and the optimizer function set to Stochastic Gradient Descent (SGD) with a learning rate of 0.2. For the CNN model, 100 filters were used with a kernel size of 3, along with the Rectified Linear Unit (ReLU) activation function. The RNN, LSTM, Bi-LSTM, GRU, and Bi-GRU models each employed 64 units. The TCN model was configured with 16 filters, a kernel size of 5, and dilation rates of [1, 2, 4, 8]. The Transformer model was set up with 2 attention heads, a hidden layer size of 64 in the feed-forward network, and the ReLU activation function. These parameter settings and architectural choices were designed to allow for a standardized comparison of the deep learning models on the IMDB dataset. This standardization facilitates an accurate analysis of each model's performance, enabling a comparison of their accuracy and loss values.

Table 2 shows the result of different deep learning models on IMDB review dataset based on various metrics including Accuracy, Precision, Recall, F1-Score, and Time of training.

**Table 2**: Result of different deep learning models on the IMDB dataset

| model | Accuracy % | Precision % | Recall % | F1-Score % | Time (h:m:s) |
|-------|-----------|-------------|----------|------------|--------------|
| CNN | 85.90 | 85.89 | 85.88 | 85.89 | 0:02:57 |
| RNN | 59.03 | 59.03 | 59.02 | 59.03 | 0:12:23 |
| LSTM | 87.53 | 87.53 | 87.54 | 87.54 | 0:09:09 |
| Bi-LSTM | 87.45 | 87.46 | 87.47 | 87.46 | 0:10:43 |
| GRU | 87.55 | 87.56 | 87.57 | 87.56 | 0:05:10 |
| BI-GRU | 87.97 | 87.92 | 87.99 | 87.95 | 0:09:54 |
| TCN | 84.42 | 84.40 | 84.42 | 84.41 | 0:07:38 |
| Transformer | **88.03** | **88.04** | **88.01** | **88.03** | 0:03:44 |

To compare the performance of these models, we utilized accuracy, validation-accuracy, loss, and validation-loss diagrams. These diagrams provide insights into how well the models are learning from the data and help in evaluating their effectiveness for sentiment classification tasks.

Fig. 28 shows the accuracy and validation-accuracy diagrams where the accuracy, provides a visual representation of how the different deep learning models perform in terms of accuracy during the training process and validation-accuracy shows the trend of accuracy values on the testing set across multiple epochs for each model.

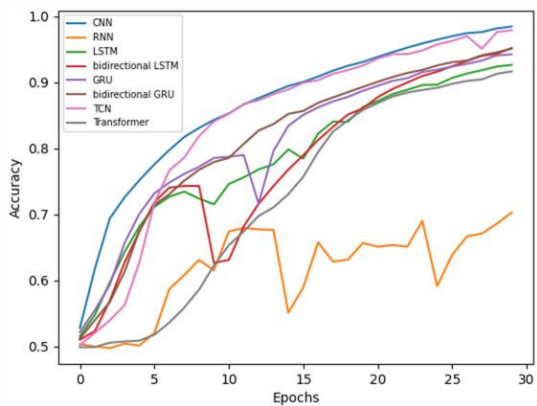
(a) Accuracy Diagram

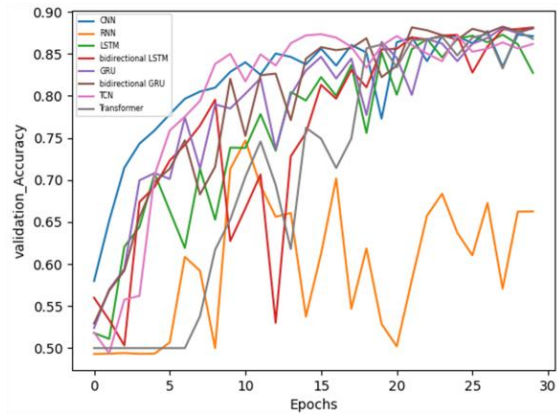
(b) Validation-Accuracy Diagram

**Figure 28.**   Accuracy and validation-accuracy of deep learning models on IMDB dataset.

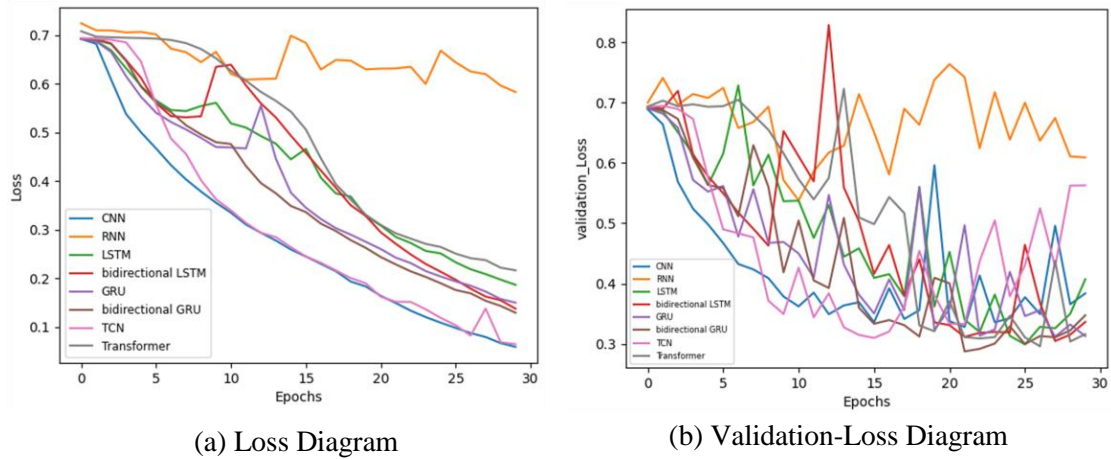

(a) Loss Diagram                    (b) Validation-Loss Diagram

**Figure 29.** Loss and validation- loss diagrams of deep learning models on IMDB dataset.

Fig. 29 illustrates the loss and validation-loss diagram where the loss diagram is a visual representation of loss values during the training process for six different models, and the validation-loss diagram depicts the variation in loss values on the testing set during the evaluation process for the different models. The loss function measures the discrepancy between the predicted sentiment labels and the actual labels.

Furthermore, the confusion matrices for the various deep learning models are displayed in Fig. 30. These matrices provide a detailed breakdown of each model's performance, highlighting how well the models classify different classes. By closely examining these confusion matrices, we can gain insights into the precision of the models and identify patterns of misclassification for each class. This analysis helps in understanding the strengths and weaknesses of the models' predictions.

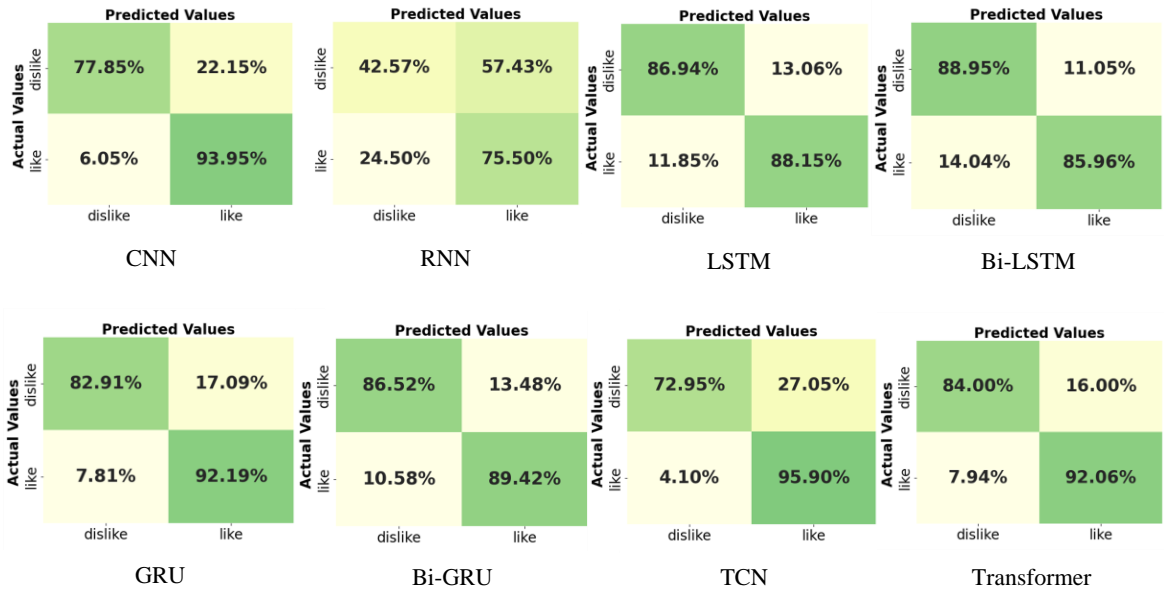

**Figure 30**. Confusion matrix for different deep learning models on IMDB dataset.

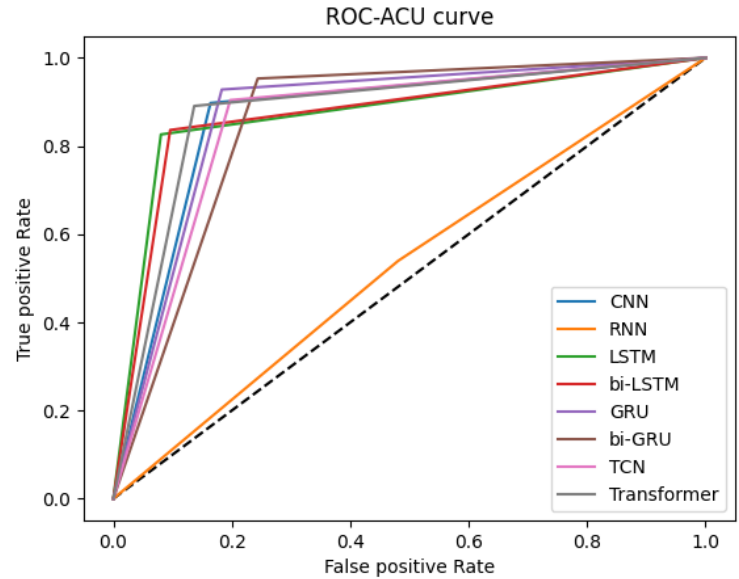

**Figure 31**. ROC-AUC diagrams for different deep learning models.

Additionally, Fig. 31 displays the ROC-AUC (Receiver Operating Characteristic-Area Under Curve) diagrams for eight different deep learning models. These diagrams offer valuable insights into the classification performance of the models, aiding in the assessment of their effectiveness. By analyzing the ROC-AUC curves, we can make informed decisions regarding model selection and threshold adjustments, ensuring a more accurate and effective classification approach.

Based on the results provided, it can be concluded that the Transformer and Bi-GRU models achieved the best performance on the IMDB review dataset for sentiment analysis. Both models demonstrated high accuracy in classifying the sentiment of movie reviews. However, it is worth noting that the training time of the Transformer model was significantly less than that of the Bi-GRU model. This suggests that the Transformer model was faster to train compared to the Bi-GRU model while still achieving excellent performance. Additionally, the GRU model also exhibited good accuracy in sentiment classification and required less training time than the Bi-GRU model. Overall, the results suggest that the Transformer, and GRU models are effective deep learning models for sentiment analysis on the IMDB review dataset, with varying trade-offs between performance and training time.

### 11.2 Methodology and Experiments on ARAS Dataset

Based on the provided information, the ARAS dataset [301] is a valuable resource for recognizing human activities in smart environments. It consists of data streams collected from two houses over a period of 60 days, with 20 binary sensors installed to monitor resident activity. The dataset includes information on 27 different activities performed by two residents, and the sensor events are recorded on a per-second basis.

Eight distinct deep learning models were used in our investigation to recognize human activities: CNN, RNN, LSTM, Bidirectional LSTM, GRU, Bidirectional GRU, TCN, and Transformer. A structural overview of the deep learning model designed to analyze the performance of eight different models on the ARAS dataset is shown in Fig. 32.

The first phase involves preprocessing the sensor data to ensure it is in a suitable and standardized format for deep learning models. The initial task in this phase is data cleaning, where any recorded instances where all sensor events are zero, and the resident is inside the house, are removed from the dataset. Next, a time-based static sliding window technique is applied for segmenting sensor events. This method groups sequences of sensor events into intervals of equal duration. Optimizing the time interval is crucial for effective segmentation; after evaluating intervals ranging from 30 to 360 seconds, a 90-second interval was determined to be optimal for the ARAS dataset. The segmentation task aids in decreasing training time and increasing accuracy for the deep learning models.

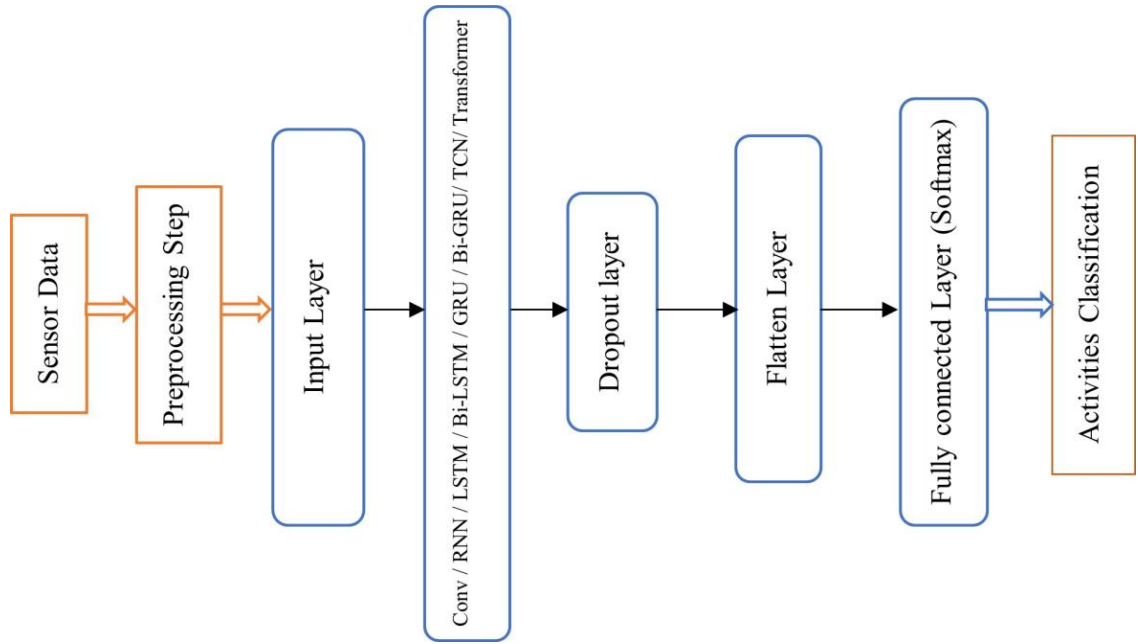

**Figure 32.** The structural for analysis of different deep learning models on the ARAS dataset

After preprocessing, the data is passed through an input layer. In the second layer, one of eight models: CNN, RNN, LSTM, Bi-LSTM, GRU, Bi-GRU, TCN, or Transformer is employed for feature extraction and training. This layer plays a vital role in capturing patterns and dependencies within the data. To mitigate overfitting, a dropout layer follows, which randomly deactivates a portion of the neurons during training, thereby improving the model's generalization. Subsequently, a flatten layer is used to convert the multi-dimensional vector into a one-dimensional vector, making it compatible with fully connected layers. Finally, the output passes through a fully connected (Dense) layer, which uses a Softmax function for classification, transforming the model's predictions into probability distributions across the classes.

In the experimental phase, we split the data from the first resident of house B, allocating 70% for training and 30% for testing, using a random split. Additionally, 20% of the training data was set aside for validation. The models were trained with a fixed set of parameters: 30 epochs, a batch size of 64, a dropout rate of 0.2, the "Categorical Crossentropy" loss function, and the Adam optimizer. For the CNN model, we used 100 filters with a kernel size of 3 and the rectified linear unit (ReLU) activation function. The RNN, LSTM, Bi-LSTM, GRU, and Bi-GRU models were configured with 64 units each. The TCN model was set with 16 filters, a kernel size of 5, and dilation rates of [1, 2, 4, 8]. The Transformer model utilized 2 attention heads, a hidden layer size of 64 in the feedforward network, and the ReLU activation function.

Table 3 illustrates the results of experiments on ARAS dataset with various metrices including Accuracy, Precision, Recall, F1-Score, and Time of training.

**Table 3**: Result of different deep learning models on the ARAS dataset

| model | Accuracy % | Precision % | Recall % | F1-Score % | Time (h:m:s) |
|---|---|---|---|---|---|
| CNN | 93.14 | 95.59 | 92.43 | 93.98 | 0:01:18 |
| RNN | 93.17 | 96.19 | 91.67 | 93.88 | 0:04:09 |
| LSTM | 93.29 | 95.56 | 92.82 | 93.81 | 0:03:23 |
| Bi-LSTM | 93.33 | **96.66** | 92.12 | 94.15 | 0:04:01 |
| GRU | 93.65 | 96.08 | 91.78 | 94.31 | 0:03:15 |
| BI-GRU | 93.90 | 95.87 | 92.61 | 94.49 | 0:03:56 |
| TCN | 94.04 | 95.37 | 93.48 | 94.42 | 0:04:06 |
| Transformer | **94.56** | 95.61 | **94.06** | **94.83** | 0:03:14 |

Also, Fig. 33 presents the accuracy diagram and validation-accuracy diagram for the deep learning models, while Fig. 34 shows the loss diagram and validation-loss diagram for deep learning models.

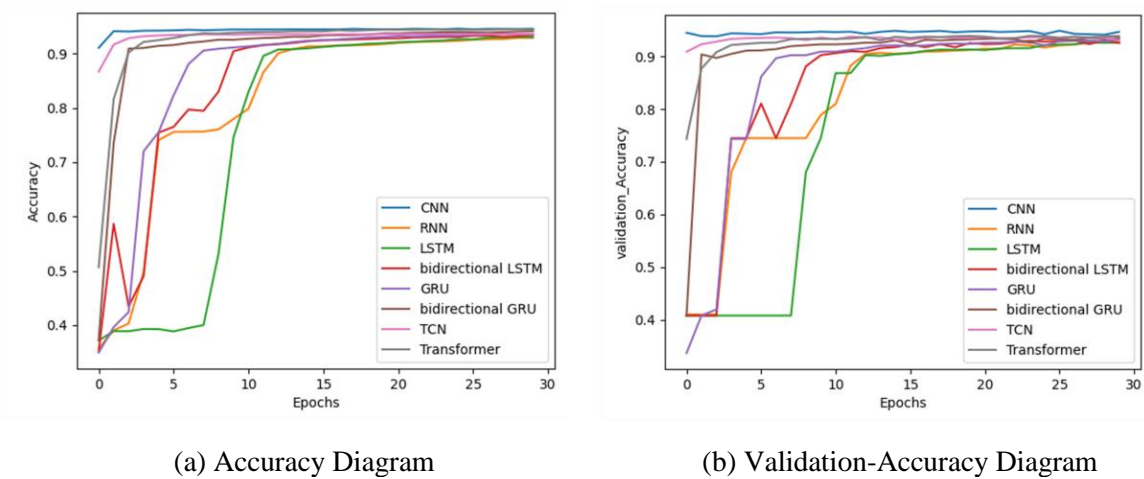

(a) Accuracy Diagram                                    (b) Validation-Accuracy Diagram

**Figure 33.** Accuracy and validation- accuracy diagrams of deep learning models on ARAS dataset**.**

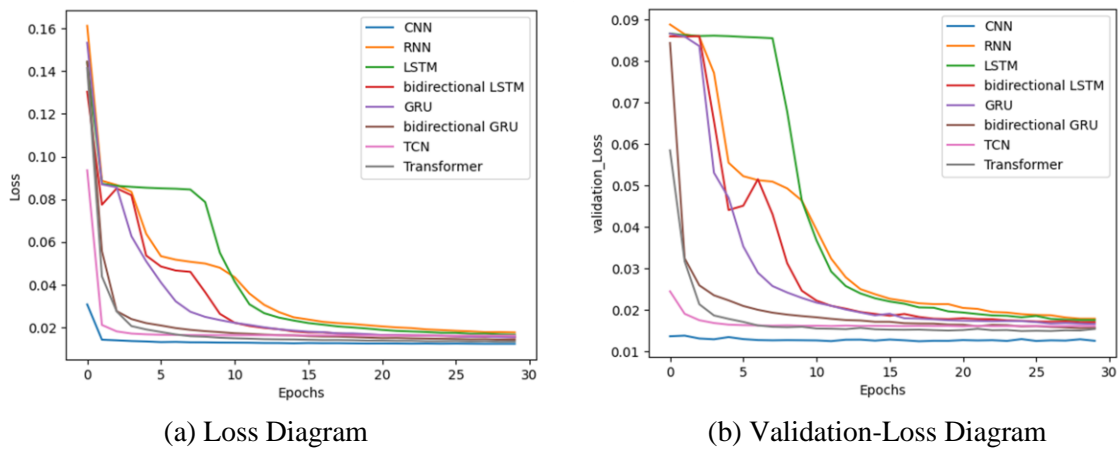

(a) Loss Diagram                                              (b) Validation-Loss Diagram

**Figure 34.**    Loss and validation- loss diagrams of deep learning models on ARAS dataset

Since we performed preprocessing tasks like data cleaning and segmentation, the data is nearly normalized and balanced, leading to consistent and closely grouped results across all models. However, the results indicate that the Transformer and TCN models outperformed the others on the ARAS dataset. This outcome aligns with the dataset's nature, which comprises spatial and temporal sequences of sensor events. Among the models, the Transformer exhibited the highest performance in terms of accuracy, recall, and F1-score, while the Bi-LSTM model excelled in the precision metric. Moreover, the Transformer model demonstrated a notable advantage in training time, second only to the CNN model, underscoring its efficiency in processing and learning from time-series data. Additionally, when examining the accuracy and loss curves, it is evident that the Transformer, TCN, and CNN models stabilized earlier than the others. Overall, the Transformer model proved to be the most effective for working with the ARAS dataset, striking a balance between accuracy, training time, and consistency throughout the training phases, making it the optimal choice for recognizing human activities based on sensor data.

### 11.3 Methodology and Experiments on the Fruit-360 Dataset

Since images are not sequential or time-dependent, recurrent models were less effective for these tasks. CNN-based models, on the other hand, are highly valuable for image analysis due to their ability to capture spatial relationships. Consequently, the analysis of deep learning models on the Fruit-360 dataset for image classification focused on eight CNN variants: VGG, Inception, ResNet, InceptionResNet, Xception, MobileNet, DenseNet, and NASNet. These models use deep transfer learning technique for training image data and improving classification accuracy. Fig. 35 provides a structural overview of the deep learning models used to evaluate the performance of these eight variants on the Fruit-360 dataset.

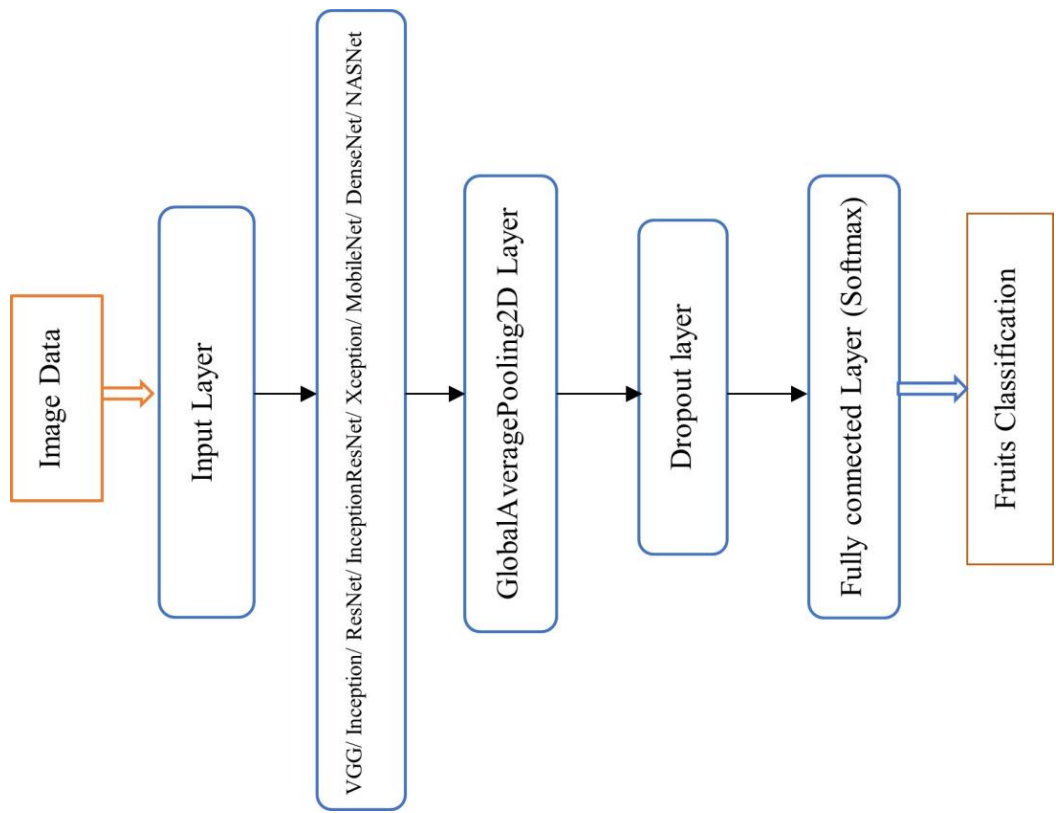

**Figure 35.** The structural for analysis of different CNN-based models on Fruit-360 dataset.

First, the fruit images are passed through an input layer. In the second layer, one of eight models (VGG, Inception, ResNet, InceptionResNet, Xception, MobileNet, DenseNet, or NASNet) is employed for feature extraction and training. Next, a Global Average Pooling 2D (GAP) layer is applied, which significantly reduces the spatial dimensions of the data by collapsing each feature map into a single value. To combat overfitting, a dropout layer is then introduced, randomly deactivating a portion of the neurons during training, which enhances the model's ability to generalize. Finally, the output is passed through a fully connected (Dense) layer, where a Softmax function is used to classify the fruit images.

The dataset comprises 55,244 images of 81 different fruit classes, each with a resolution of $100 \times 100$ pixels. For the experiments, a subset of 60 fruit classes was selected, containing 28,484 images for training and 9,558 images for testing. Non-fruit items such as chestnuts and ginger root were removed from the dataset.

All models were trained with a consistent set of parameters: 20 epochs, a batch size of 512, a dropout rate of 0.2, the "Categorical Crossentropy" loss function, and the Adam optimizer. Additionally, all models utilized the "ImageNet" dataset for pre-training.

Table 4 presents the experimental results for various models on the Fruit-360 dataset, including VGG16, InceptionV3, ResNet50, InceptionResNetV2, Xception, MobileNet, DenseNet121, and NASNetLarge. The table includes metrics such as Accuracy, Precision, Recall, F1-Score, and Time of training.

**Table 4:** Result of different deep learning models on the Fruit360 dataset

| model | Accuracy % | Precision % | Recall % | F1-Score % | Time (h:m:s) |
|---|---|---|---|---|---|
| VGG | 94.39 | **99.79** | 80.65 | 89.20 | 2:17:32 |
| Inception | 95.86 | 96.65 | 95.14 | 95.89 | 0:23:34 |
| ResNet | 94.59 | 95.30 | 93.64 | 94.46 | 1:12:56 |
| InceptionResNet | 96.05 | 97.01 | 95.36 | 96.18 | 0:54:18 |
| Xception | 97.38 | 98.28 | 96.61 | 97.44 | 1:01:11 |
| MobileNet | 98.54 | 98.88 | 98.28 | 98.58 | 0:17:22 |
| DenseNet | **98.94** | 99.12 | **98.75** | **98.94** | 1:10:30 |
| NASNet | 96.99 | 97.69 | 96.56 | 97.12 | 3:50:05 |

Furthermore, the accuracy, validation-accuracy, loss, and validation-loss diagrams were used to compare the performance of various models. When assessing the models' performance for tasks involving the categorization of fruit photos, these graphs offer valuable insights into how effectively the models are learning from the data. Fig. 36 shows the accuracy and validation-accuracy diagram of the deep learning models, while Fig. 37 illustrates the loss diagram and validation-loss diagram of the deep learning models.

Based on the results, it can be concluded that the DenseNet and MobileNet models achieved the best performance for fruit image classification on the Fruit-360 dataset. Both models demonstrated high accuracy in classifying fruit images. Notably, MobileNet had a significantly shorter training time compared to DenseNet, indicating that it was faster to train while still delivering performance close to that of DenseNet. Additionally, the Xception model also showed good accuracy and required less training time than DenseNet. Overall, the MobileNet model stands out as a favorable choice due to its balance between accuracy and training efficiency.

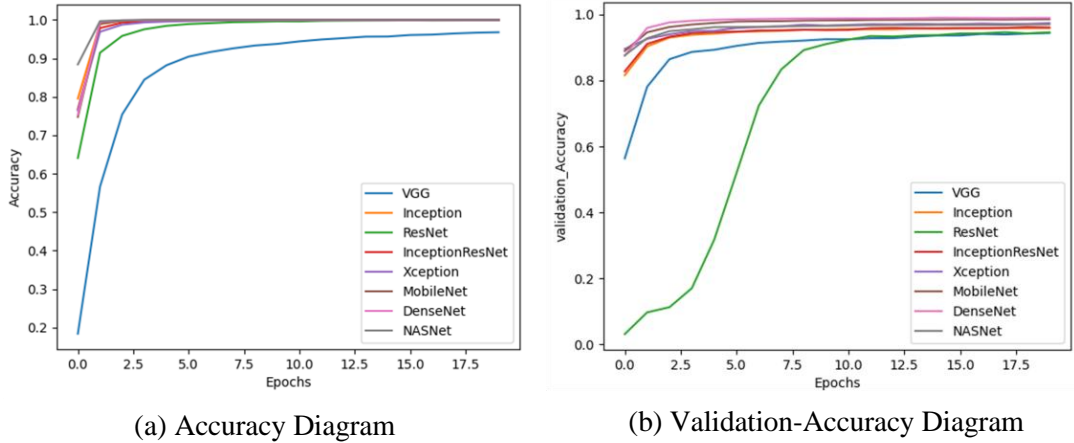

(a) Accuracy Diagram                              (b) Validation-Accuracy Diagram

**Figure 36.**   Accuracy and validation- accuracy diagrams of different CNN-based deep learning models on Friut-360 dataset.

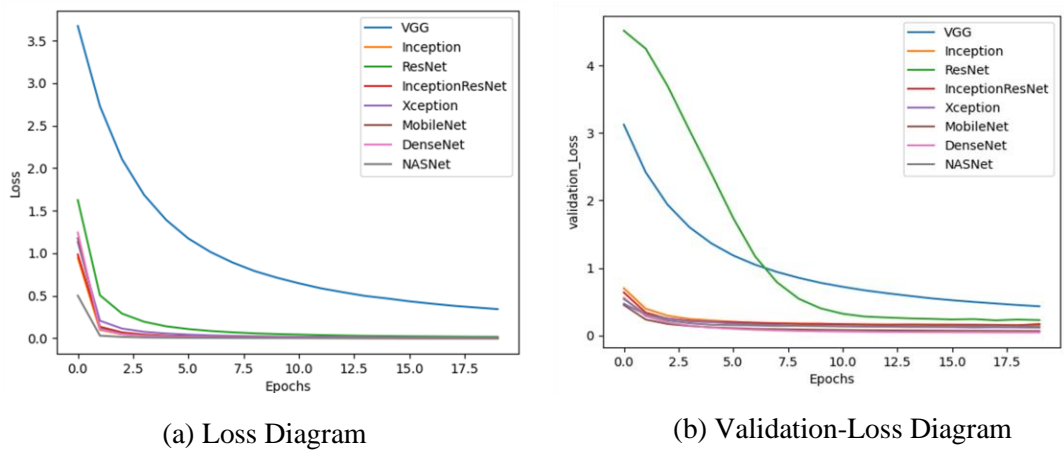

(a) Loss Diagram                                  (b) Validation-Loss Diagram

**Figure 37.**   Loss and validation- loss diagrams of different CNN-based deep learning models on Friut-360 dataset.

## 12 Research Directions and Future Aspects

In the preceding sections, we explored a range of deep learning topics, highlighting both the advantages and limitations of various deep learning models. Additionally, we examined the application of several models across different domains. Despite the benefits demonstrated, our research has identified certain gaps, indicating that further advancements are necessary. This section outlines potential future research directions based on our analysis.

- *Generative (Unsupervised) Models:* Generative models, a key category of deep learning models discussed in Section 4, hold significant promise for future research. These models enable the creation of new data representations through exploratory analysis and can identify high-order correlations or features in data. Unlike supervised learning, unsupervised models can derive insights from data without the need for labeled examples, making them valuable for various applications. Several generative models, including Autoencoders, Generative Adversarial Networks (GANs), Deep Belief Networks (DBNs), and Self-Organizing Maps (SOMs), have been developed and employed across diverse contexts. A promising research avenue involves analyzing these models in various settings and developing new methods or variations that enhance data modeling or representation for specific real-world applications. The rising interest in GANs is particularly noteworthy, as they excel in leveraging unlabeled image data for deep representation learning and training highly non-linear mappings between latent and data spaces. The GAN framework offers the flexibility to formulate new theories and methods tailored to emerging deep learning applications, positioning it as a pivotal area for future exploration.

-   *Hybrid/Ensemble Modeling*: Hybrid deep learning architectures have shown great potential in enhancing model performance by combining components from multiple models. For instance, the integration of Convolutional Neural Networks (CNNs) and Recurrent Neural Networks (RNNs) can capture both temporal and spatial dependencies in data, leveraging the strengths of each model. Hybrid models also benefit from combining generative and supervised learning, offering superior performance and improved uncertainty handling in high-risk scenarios. Developing effective hybrid models, whether supervised or unsupervised, presents a significant research opportunity to address a wide range of real-world problems, including semi-supervised learning tasks and model uncertainty. This approach moves beyond conventional, isolated models, emphasizing the need for sophisticated methods that can handle the complexity of various data types and applications.

-   *Hyperparameter Optimization for Efficient Deep Learning*: As deep learning models have evolved, the number of parameters, computational latency, and resource requirements have increased substantially [152]. Selecting the appropriate hyperparameters is critical to building a neural network with high accuracy. Key hyperparameters include learning rate, loss function, batch size, number of training iterations, and dropout rate, among others. The challenge lies in finding an optimal balance of these parameters, as they significantly influence network performance. However, iterating through all possible combinations of hyperparameters is computationally expensive. To address this, metaheuristic optimization techniques, such as Genetic Algorithm (GA) [303], Particle Swarm Optimization (PSO) [304], and others, can be employed to explore the search space more efficiently than exhaustive methods. Future research should focus on optimizing hyperparameters tailored to specific data types and contexts. For example, the learning rate plays a crucial role in training, where a rate too high may cause the model to converge prematurely, while a rate too low can lead to slow convergence and prolonged training times. Adaptive learning rate techniques, such as including Adaptive Moment Estimation (Adam) [305], Stochastic Gradient Descent (SGD) [306], adaptive gradient algorithm (ADAGRAD) [307], and Nesterov-accelerated Adaptive Moment Estimation (Nadam) [308], and more recent innovations like Evolved Sign Momentum (Lion) [309], offer promising avenues for improving network performance and minimizing loss functions. Future research could further explore these optimizers, focusing on their comparative effectiveness in enhancing model performance through iterative weight and bias adjustments.

-   *Federated Learning*: Federated learning is an emerging deep learning paradigm that enables collaborative model training across multiple organizations or teams without the need to share raw data. This approach is particularly relevant in contexts where data privacy is paramount. However, federated learning introduces new challenges, especially with the advent of data fusion technologies that combine data from multiple sources with varying formats. As data diversity and volume continue to grow, optimizing data and model utilization in federated learning becomes increasingly important. Addressing challenges such as safeguarding user privacy, developing universal models, and ensuring the stability of data fusion outcomes will be crucial for the future application of federated learning across multiple domains [310].

-   *Quantum Deep Learning:* Quantum computing and deep learning have both seen significant advancements over the past few decades. Quantum computing, which leverages the principles of quantum mechanics to store and process information, has the potential to outperform classical supercomputers on certain tasks, making it a powerful tool for complex problem-solving. The intersection of quantum computing and deep learning has led to the emergence of quantum deep learning and quantum-inspired deep learning algorithms. Future research directions in this area include investigating and developing quantum deep learning models, such as Quantum Convolutional Neural Network (Quantum CNN) [311], Quantum Recurrent Neural Network (Quantum RNN) [312], Quantum Generative Adversarial Network (Quantum GAN) [313], and others. Additionally, exploring the application of these models across various domains and creating novel quantum deep learning architectures represents a cutting-edge frontier in the field [314, 315].

In conclusion, the research directions outlined above underscore the dynamic and evolving nature of deep learning. By addressing these challenges and exploring new avenues, the field can continue to advance, driving innovation and enabling the development of more powerful and efficient models for a wide range of applications.

## 13 Conclusion

This article provides an extensive overview of deep learning technology and its applications in machine learning and artificial intelligence. The article covers various aspects of deep learning, including neural networks, MLP models, and different types of deep learning models such as CNN, RNN, TCN, Transformer, generative models, DRL, and transfer learning. The classification of deep learning models allows for a better understanding of their specific applications and characteristics. The RNN models, including LSTM, Bi-LSTM, GRU, and Bi-GRU, are particularly suited for time series data due to their ability to capture temporal dependencies. On the other hand, CNN-based models excel in image data analysis by effectively capturing spatial features.

The experiments conducted on three public datasets, namely IMDB, ARAS, and Fruit-360, further reinforce the suitability of specific deep learning models for different data types. The results demonstrate that the CNN-based models such as DenseNet and MobileNet perform exceptionally well in image classification tasks. The RNN models, such as LSTM and GRU, show strong performance in time series analysis. However, the Transformer model outperforms classical RNN-based models, particularly in text analysis, due to its use of the attention mechanism.

Overall, this article highlights the diverse applications and effectiveness of deep learning models in various domains. It emphasizes the importance of selecting the appropriate deep learning model based on the nature of the data and the task at hand. The insights gained from the experiments contribute to a better understanding of the strengths and weaknesses of different deep learning models, facilitating informed decision-making in practical applications.

**Acknowledgement:** The authors would like to express sincere gratitude to all the individuals who have contributed to the completion of this research paper. Their unwavering support, valuable insights, and encouragement have been instrumental in making this endeavor a success.

**Funding Statement:** The authors received no specific funding for this study.

**Author Contributions:** The authors confirm contribution to the paper as follows: Study conception and design: F. M. Shiri, T. Perumal; data collection: F. M. Shiri; analysis and interpretation of results: F. M. Shiri, T. Perumal, N. Mustapha, R. Mohamed; draft manuscript preparation: F. M. Shiri, T. Perumal, N. Mustapha, R. Mohamed. All authors reviewed the results and approved the final version of the manuscript.

**Availability of Data and Materials:** The code used and/or analyzed during this research are available from the corresponding author upon reasonable request. Data used in this study can be accessed via the following links:

IMDB dataset: https://ai.stanford.edu/~amaas/data/sentiment/, 6/19/2011

ARAS dataset: http://aras.cmpe.boun.edu.tr/download.php, 7/22/2013

Fruit360 dataset: https://data.mendeley.com/datasets/rp73yg93n8/1, 10/20/2018

**Conflicts of Interest:** The authors declare that they have no conflicts of interest to report regarding the present study.

**Ethics Approval:** Not applicable.